\documentclass[runningheads]{llncs}

\usepackage[pdftex]{hyperref}
\usepackage{times}
\usepackage{comment}
\usepackage[utf8]{inputenc}
\usepackage[numbers]{natbib}
\setcitestyle{square}
\usepackage{mathptmx}
  
\usepackage{array}
\usepackage{amsfonts}
\usepackage{amsmath}
\usepackage{amssymb}
\usepackage{bm}
\usepackage{enumerate}
\usepackage{graphics}
\usepackage{graphicx}
\usepackage{epstopdf}
\usepackage{pdflscape}
\usepackage{lineno}
\usepackage{doi}
\usepackage[boxruled,linesnumbered]{algorithm2e}
\usepackage[boxruled,linesnumbered]{algorithm2e}
\usepackage{pythonhighlight}

\newcommand{\sectref}[1]{Sect.~\ref{#1}}
\newcommand{\algref}[1]{Algorithm~\ref{#1}}

\newcommand{\figref}[1]{Fig.~\ref{#1}}

\newcommand{\myvec}[1]{{\bm {#1}}}  % For vectors (Elsevier)
\DeclareMathOperator{\sign}{sign}

\newcounter{ex}
\newcommand{\qreff}[2]{{\bf Example\refstepcounter{ex} \arabic{ex}
		\def\@currentlabel{\arabic{ex}} \label{#2}} {(#1)}}
\usepackage{graphicx} % Required for inserting images

\begin{document}

\title{A Programmer's Guide to Cascaded Adaptive Combiners: Online Learning by Biologically Accurate Models of Multilayer Neuron Networks}

\titlerunning{Programmer's Guide to Cascaded Adaptive Combiners}

\author{Martin Nilsson\inst{1} \and  Denis Kleyko\inst{2,1}}

\authorrunning{M. Nilsson and D. Kleyko}

\institute{RISE Research Institutes of Sweden, Kista, Sweden
\and
\"Orebro University, \"Orebro, Sweden
}

\maketitle

\begin{abstract}
Learning in biological multilayer neuronal networks offers insights that extend beyond the classical weighted-sum neuron model commonly used in artificial neural networks. This article presents an accessible guide to a mechanistic neuronal network model that more accurately captures aspects of biological computation while enabling a simple yet powerful mechanism for learning in multilayer neural networks. The proposed approach supports efficient online streamed learning and provides a practical alternative to backpropagation. We demonstrate its potential in an image classification task, achieving competitive classification performance.
The approach's simplicity, biological grounding, and broad applicability highlight a promising path toward algorithms that unify mechanistic neuron models and machine learning.
\end{abstract}


\section{Introduction}
\subsection*{The classical artificial neuron model is oversimplified}

The classical neuron model used in artificial neural networks consists of a weighted sum 
followed by the application of an activation function~\cite{Rosenblatt.1958tpa}. 
This formulation is mathematically convenient and has proven highly effective in many practical applications.
However, when compared to biological neurons, it represents an extreme simplification.
More detailed neuron models have been developed within computational neuroscience, but these are often sufficiently complex, making large-scale implementation challenging ~\cite{izhikevich2003simple}. 
Moreover, many such models are primarily {\em empirical} in nature and are not always designed to provide predictive insight into how biological neurons operate within large networks. 
In particular, plasticity mechanisms are frequently introduced phenomenologically rather than derived from an explicit account of underlying cellular processes.

A useful distinction can be made between empirical and {\em mechanistic} models. 
Empirical models are typically constructed to reproduce observed data without necessarily modeling the internal processes of the neuron.
This approach can yield excellent agreement with experiments. 
However, it may offer limited guidance when extrapolating to untested conditions or when attempting to abstract principles to higher organizational levels. 
By contrast, mechanistic models aim to capture the internal operations of the neuron, providing a constructive explanation of its behavior~\cite{Craver.2006wmm}. 
Although such models may be more difficult to fit precisely to experimental data, they can offer greater interpretability and potentially stronger predictive power on how the neuron would behave in new situations.
They also facilitate mathematical abstraction and the study of collective dynamics in large neural systems.

Why emphasize mechanistic modeling? Biological systems demonstrate highly efficient and robust learning capabilities. 
Understanding the mechanisms underlying these capabilities may provide a principled foundation for developing intelligent systems with similar efficiency and generality. 
Biologically inspired approaches can certainly yield practical successes, but grounding them in mechanistic models may offer a more systematic path toward robust and general learning architectures.

\subsection*{A more realistic neuron model: The adaptive conical combiner}
In this study, we adopt a neuron model that departs from the classical formulation while remaining computationally tractable.  
The model is grounded in a mathematical abstraction of established neurobiological properties of ion channels and their signal processing function in neurons. 
It has been described in several previous studies~\cite{Nilsson.Jorntell.2021ccf,Nilsson.2023ipb,Nilsson.2023meo}.

A key property of this model is the separation of inputs into inhibitory and excitatory ones. 
Only the excitatory inputs possess variable weights, whereas inhibitory weights are fixed. 
This division reflects principles widely discussed in neurophysiology, including those associated with Gray's rules 
\citep{Gray.1959asa,Harris.Weinberg.2012uos}. 
Additionally, the model constrains synaptic weights to be non-negative.

In contrast, the classical artificial neuron does not explicitly distinguish between inhibitory and excitatory inputs, instead allowing signed weights. 
While this choice simplifies mathematical treatment and optimization, it abstracts away certain structural constraints that may play functional roles in biological systems.
As argued in~\cite{Nilsson.2023ipb}, enforcing such constraints can enrich representational properties, including the organization of hierarchical structure.

\subsection*{Plasticity and learning}

An essential property of neurons is \emph{plasticity}, the ability to modify their responses based on experience. 
In biological systems, plasticity often involves local feedback  circuit, and this loop crucially depends on one set of inputs having fixed weights while the other has variable weights. 
In the adopted model, the separation between fixed inhibitory inputs and adaptive excitatory inputs enables the neuron to function as an {\em adaptive combiner}, a well-established concept in signal processing. 
Fundamentally, an adaptive combiner operates by adjusting the weights for the excitatory inputs (the ``feature vector'') to approximate inhibitory inputs (the ``reference signal''), and the neuron's output corresponds to the resulting approximation error. 
When many such processing elements are arranged in parallel layers, the resulting computations can be highly expressive~\cite{Nilsson.2023ipb}.

The classical neuron model does not inherently specify a particular plasticity mechanism.
In artificial neural networks, learning is typically achieved through externally defined optimization procedures. 
The most widely used method is backpropagation~\citep{Baydin.et.al.2018adi}, in which weight updates depend on gradients computed from the network's output error. 
Although highly effective in applications, this procedure relies on assumptions, such as global error signals, local storage of complex information, and precise weight transport that do not have direct counterparts in biological circuits~\citep{lillicrap2020backpropagation}. 
Consequently, considerable research has explored biologically motivated alternatives and approximations~\citep{zenke2021brain}.

This raises the broader question of how biological neurons achieve efficient learning under the constraints of local computation. 
Experimental evidence indicates that neurons can adapt rapidly and operate on signals with high temporal precision, suggesting that local feedback mechanisms play a central role. 
From a signal processing perspective, stable and efficient adaptation requires feedback loops with delays short relative to the signal dynamics for the system to remain stable, which reinforces the plausibility of locally implemented learning rules.

\subsection*{Layers of neurons and multilayer networks}

As discussed above, layers composed of similar neurons can implement powerful computations. 
The next step in studying larger systems of neurons is to consider multiple stacked layers, which we refer to as {\em cascaded adaptive combiners} (CACs).
As we will demonstrate below, CACs provide an efficient and general learning method, that avoids some limitations commonly associated with backpropagation.

\subsection*{High-level comparison of backpropagation networks and CAC}

\begin{figure}[!ht]
    \begin{minipage}{0.46\linewidth}
    \center{\includegraphics[width=1.0\linewidth]{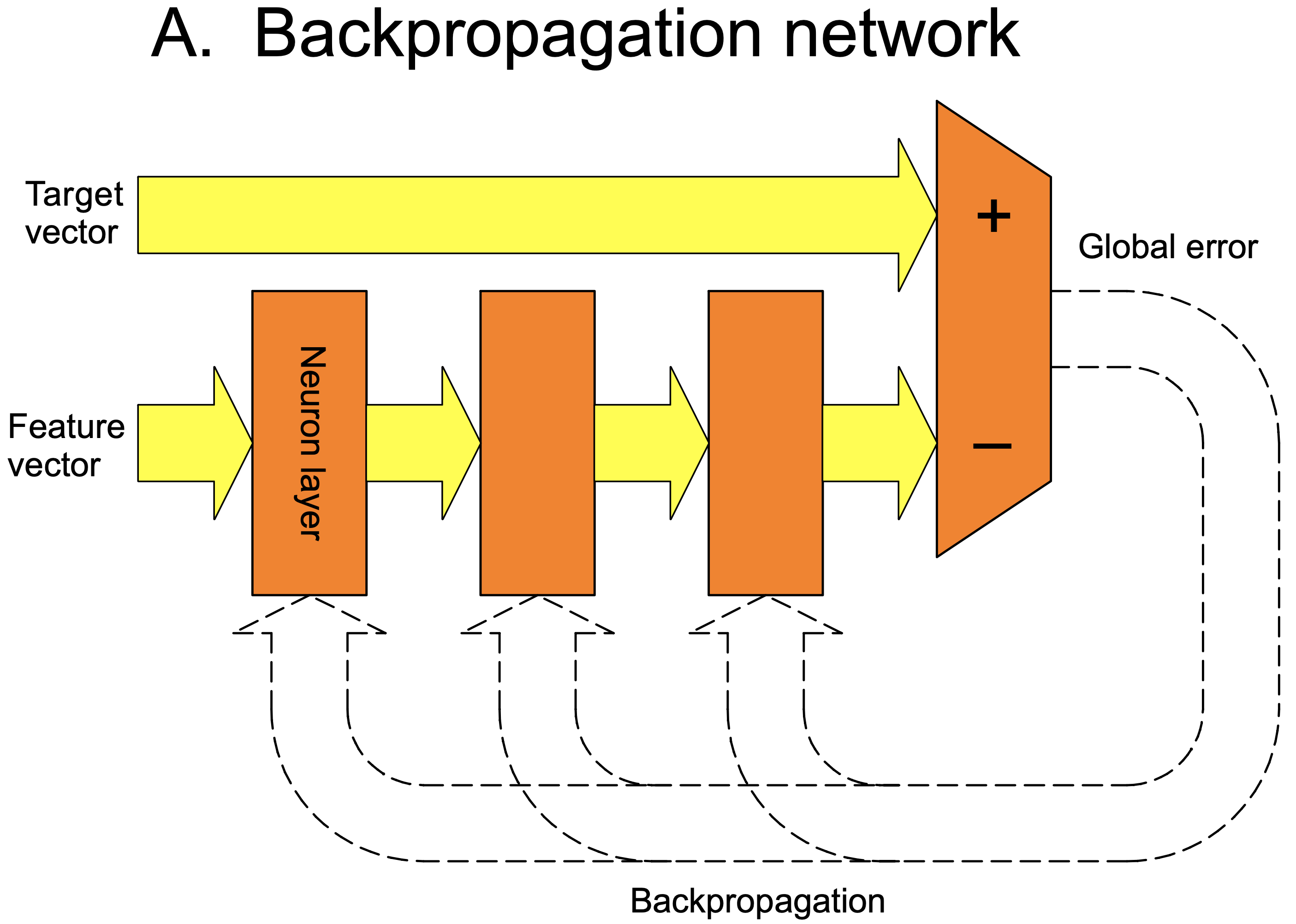}}
    \end{minipage}
    \hfill
    \begin{minipage}{0.52\linewidth}
    \center{\includegraphics[width=0.95\linewidth]{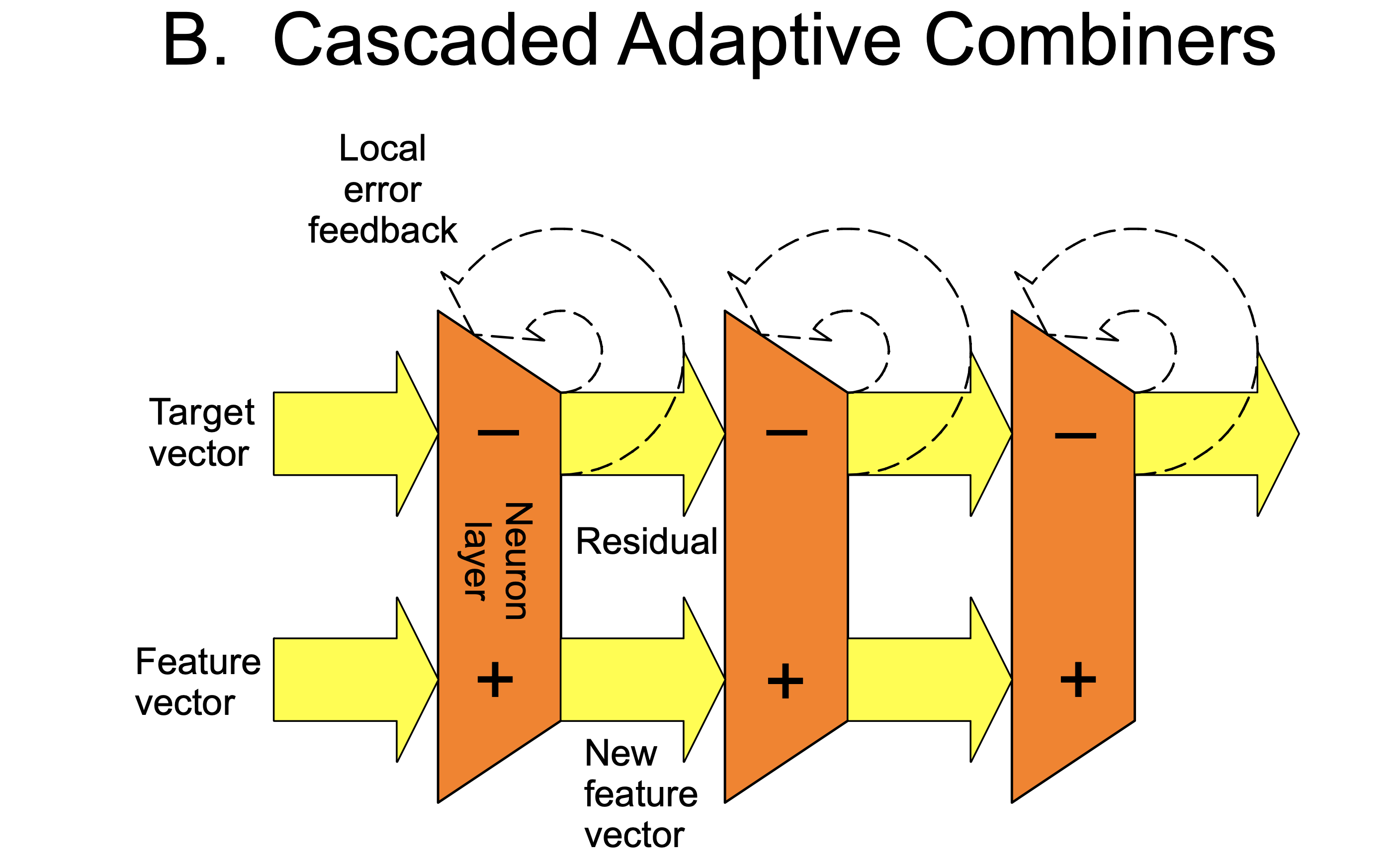}}
    \end{minipage}
    \caption{{\bf Comparison of conventional backpropagation networks with Cascaded Adaptive Combiners.} 
    {\bf A.} The left panel is a high-level visualization of a conventional neural networks with backpropagation. 
    It passes feature vectors through multiple layers of artificial neurons. At the end, the output is compared to the reference signal and the
    error is propagated back through the whole network. Due to the global feedback, this structure is not suitable for
    online learning. {\bf B.} The right panel shows the structure of the network formed by cascading adaptive combiners. In this case, the layers are differential, where each layer compares the residual of the reference signal or the error term from the previous layer with a transformation that is derived from the input feature vector. The network strives to reduce the error term to zero. Here, local error feedback is sufficient; hence, this architecture is suitable for online learning. 
    }
    \label{fig:comparison-bpn-cac}
\end{figure}

In a conventional neural network trained via backpropagation, feature vectors are passed through multiple layers of neurons, \figref{fig:comparison-bpn-cac}A. After the feed-forward pass, the output is compared with a reference signal, and the resulting error is propagated back through all layers to adjust the weights in order to minimize the difference. However, this backpropagation phase, along with the need to store local gradients, 
%prevents online learning as new data cannot be introduced while the system is training. 
increases memory usage and introduces synchronization constraints between forward and backward passes, which makes online learning challenging.
As a result, large neural networks are  %invariably trained offline.
often trained using mini-batches or in offline settings.

In contrast, CACs utilize differential elements, allowing both feature vectors and reference signals to be fed simultaneously into the first layer, \figref{fig:comparison-bpn-cac}B. They are matched locally within the same layer, and the weights are updated locally. The error, or residual, is output from the layer and used as the reference signal for the next layer. At the same time, a non-linear transformation is applied to the original feature vector, generating a new feature vector for the next layer.

CAC's learning process is a successive approximation procedure, aiming to eliminate the error term by incrementally removing predictable components, {\it i.e.}, transformations of the feature vector. The key advantage of this approach is that it does not require a global update phase, such as during backpropagation. Updates are localized to each layer and can be pipelined, making CAC highly suitable for online learning and parallel processing.
 
\subsection*{Overview of the article}
This article introduces CACs primarily from a programmer's perspective. For this reason, it focuses on algorithms and their implementation, and presents programming examples.
%Several programming examples are given, and more advanced examples are developed from simpler ones step by step.
In the next section, we introduce a class of adaptive combiners as fundamental neuron models and relate it to several classical models of historical significance. We conclude this section with the most biologically accurate of this class of models: the adaptive conical combiner. In sect.~\ref{sec:cacs}, we consider networks of multiple neurons of
gradually increasing size: first as layers of neurons and then as multiple stacked layers, CACs. 
In sect.~\ref{sec:combiners:mnist}, we present experimental results from application to the MNIST dataset \cite{Deng.2012tmd}. 
The subsequent sections discuss the approach and the results and draw some conclusions.
The reference section is followed by an appendix containing two more elaborate examples of CAC implementations.

\section{Adaptive combiners}
In this section, we introduce a single neuron as a signal processing algorithm and present an evolution of related ideas. 

\subsection{The basic adaptive combiner}
\label{sec:basic-approx}
The adaptive combiner is an iterative algorithm that accepts as input streams of scalars
$y(t)$ and vectors $\myvec{x}(t) = (x_1(t), x_2(t),\ldots, x_n(t))^\top$, where $t$ are discrete time points $t=0, 1,2,\ldots$
The combiner's job is to find a combination
of the components of $\myvec{x}$  that best approximates $y$,
{\it i.e.}, to find weights $\myvec{w} =  (w_1, w_2,\ldots, w_n)^\top$ 
so that the inner product of $\myvec{w}$ and $\myvec{x}(t)$
approximates $y(t)$, 
\begin{equation}
\label{eq:basic-approx}
    y(t) \approx \myvec{w}^\top \myvec{x}(t) =  w_1 x_1(t) + w_2 x_2(t) +\ldots + w_k x_k(t)  +\ldots + w_n x_n(t) .
\end{equation}
Thus, the adaptive combiner can be viewed  as a program that adjusts its parameters to mimic the external reference signal $y$ in terms of the components $x_k$.

The adaptive combiner is worth studying because it can perform advanced signal processing despite its simplicity. 
There are many variants of the adaptive combiner. The most fundamental adaptive combiner is the
\emph{adaptive linear combiner} (sect.~\ref{sec:linear:comb}) proposed by Widrow and Hoff in the late 1950s. Other variants
include the 
\emph{Perceptron}~\citep{McCulloch.Pitts.1943alc}, 
\emph{ADALINE}~\cite{Widrow.Hoff.1960asc}, 
\emph{the adaptive filter}~\cite{Widrow.Stearns.1985asp}, and the 
\emph{adaptive conical combiner}~\cite{Nilsson.2023meo}.
Each is introduced in detail in the following sections.

\subsection{The adaptive linear combiner}
\label{sec:linear:comb}
\begin{figure}[!ht]
	\centering{\includegraphics[width=0.7 \textwidth]{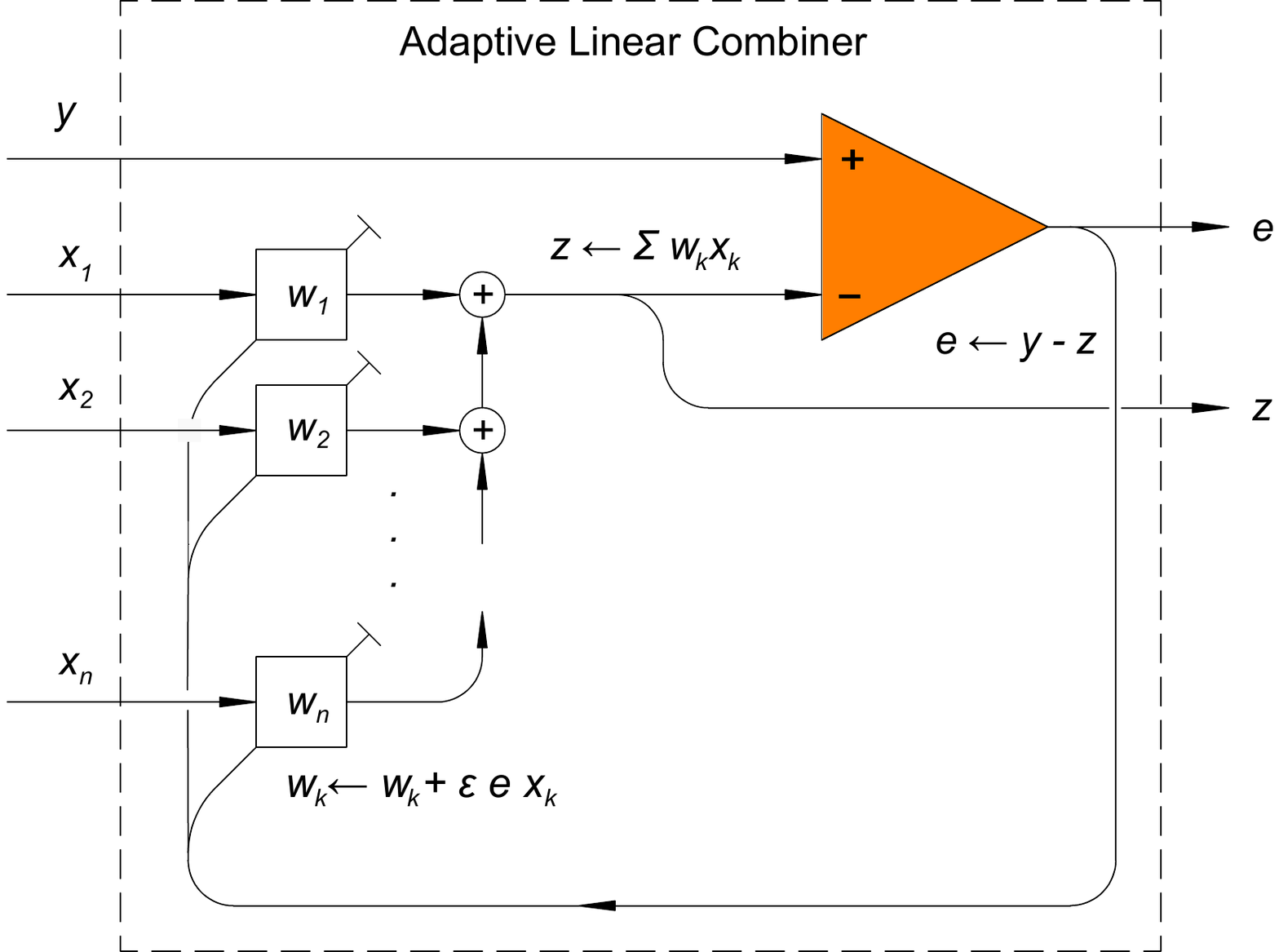}}
	\caption{{\bf The adaptive linear combiner}.
    Inputs are the reference signal $y$ and the feature vector $\myvec{x}$. The weights $\myvec{w}$ are learned with the least mean squares learning rule \protect{\eqref{eq:lms-rule}}.    
    The output is the prediction term $z$ and the error $e$. 
		The adaptive linear combiner is the most basic form of the adaptive combiner. 
        %Several variations on the theme are described below.
		\label{fig:adaptive-linear-combiner}}
\end{figure}
The principal outputs of the adaptive linear combiner (\figref{fig:adaptive-linear-combiner}) are the \emph{prediction} $z(t) = \myvec{w}^\top \myvec{x}(t)$ and the \emph{error} or \emph{residual} $e(t) = y(t) - z(t)$. 
There is a simple update rule for the weights using the
error as the feedback,
\begin{equation}
\label{eq:lms-rule}
	\myvec{w} \leftarrow \myvec{w} + \varepsilon \, e(t) \myvec{x}(t),
\end{equation}
\noindent
where $\varepsilon$ is the \emph{learning rate}.
This rule is known as the  \emph{least mean squares rule} (LMS rule), and was proposed by Widrow and Hoff~\citep{Widrow.Hoff.1960asc}. It is known by several alternative names, including the \emph{delta rule} and the \emph{covariance rule}. 
This rule has been tested extensively in analytical and applied settings and has turned out to be robust. 
It converges under quite general circumstances,  as long as the learning rate $\varepsilon$ is not too large, which follows from the fact that the LMS rule is a special case of stochastic gradient descent.
Pseudocode for the adaptive linear combiner with the LMS rule is shown in \algref{alg:adaptive-linear-combiner}.

\begin{algorithm}[!ht]
	\caption{{\bf Adaptive linear combiner}. The least mean squares rule is used to update the weights.}
    \label{alg:adaptive-linear-combiner}
	\SetKwInOut{Input}{input}
	\SetKwInOut{Constant}{constant}
	\SetKwInOut{Local}{local}
	\SetKwInOut{Output}{output}
	\SetKwComment{Comment}{/* }{ */}
	\Input{$x_k(t), 1\le k\le n$ \Comment*[r]{Component signals} }
	\Input{$y(t)$ \Comment*[r]{Reference signal}}
	\Constant{$n \ge 1$ \Comment*[r]{Number of components}}
	\Constant{$\varepsilon > 0$ \Comment*[r]{Learning rate}}
	\Output{$z(t)$  \Comment*[r]{Prediction}}
	\Output{$e(t)$  \Comment*[r]{Error}}
	\Local{$w_k, 1\le k\le n$  \Comment*[r]{Weights} }
	\For{$t \gets 0, 1,\ldots$}{   
		$z(t) \gets \Sigma_{k=1}^n w_k x_k(t)$ \Comment*[r]{Compute prediction}
		$e(t) \gets y(t) - z(t)$ \Comment*[r]{Compute error}
		\For {$k \gets 1,\ldots, n$} {
			$w_k \gets w_k + \varepsilon \, e(t) x_k(t)$ \Comment*[r]{Update weights}
		}
	}
\end{algorithm} 

%\newpage

\subsubsection{Programming}

\begin{figure}[!ht]
	\centering{\includegraphics[width=0.99 \textwidth]{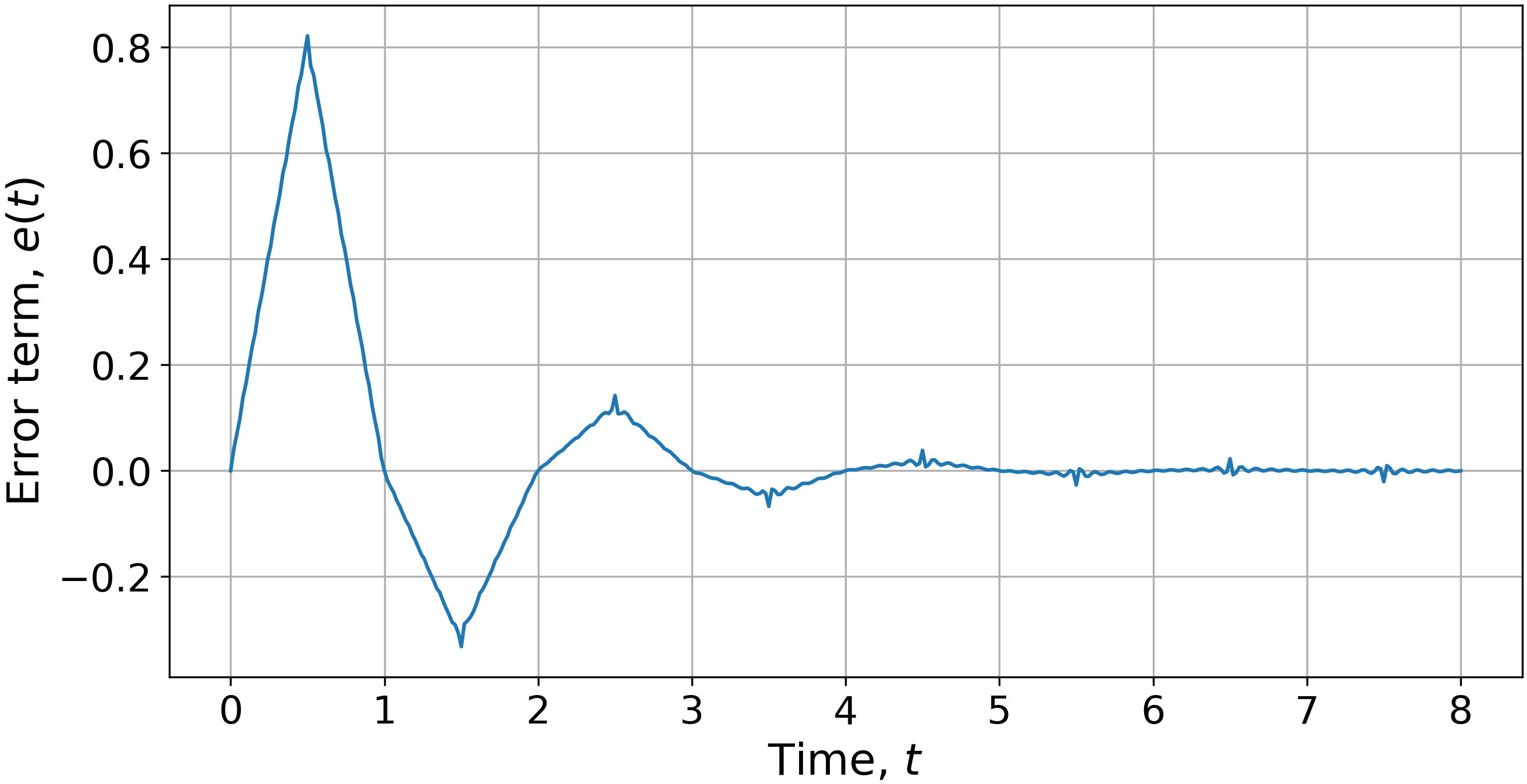}}
	\caption{{\bf The error term as a function of time for the adaptive linear combiner}, \algref{alg:adaptive-linear-combiner}.
		The plot depicts the convergence by showing how the error $e$ decays as a function of time.
        \label{fig:diagram}}
\end{figure}

The code below shows an example of the adaptive linear combiner implemented in Python. The inputs are defined as generators. As an example, we can define the $y$ input as a triangle wave, and the $x$ inputs as a set of sine waves:
\begin{python}[basicstyle=\scriptsize\ttfamily]
import numpy as np
from scipy import signal

# Sampling interval
dt = 0.02           

def ystream():
    """ Generator for the reference signal, y. """
    t = 0
    while True:
        # Triangle wave from SciPy
        yield signal.sawtooth(((t*dt+0.5)*np.pi), width = 0.5)
        t += 1

def xstream():
    """ Generator for the feature vector, x. """
    t = 0
    while True: 
        # A few sine waves
        yield np.array([np.sin(np.pi * k * t * dt) for k in range(1,20)])
        t += 1
\end{python}
\noindent
We can then implement the adaptive linear combiner itself:
\begin{python}[basicstyle=\scriptsize\ttfamily]
# Learning rate
epsilon = 0.03     

def adaptive_combiner(ys, xs):
    """ Adapt combination of x-components to y and
        generate the output streams. """

    for y, x in zip(ys, xs):

       # Initialize weights if necessary
       try: 
           w
       except: 
           w = np.zeros(len(x))

       # Compute the prediction term
       z = w.dot(x)
    
       # Compute the error term
       e = y - z
    
       # Output prediction term, error term, and weights
       yield (z, e, w)
    
       # Update weights according to the LMS rule
       w = w + epsilon * e * x
\end{python}
\noindent
Finally, we can test the implementation and plot the error with the following code:
\begin{python}[basicstyle=\scriptsize\ttfamily]
# Start up the adaptive combiner
zstream = adaptive_combiner(ystream(),xstream())

# Plot the error
import matplotlib.pyplot as plt
from itertools import islice
elist = [e for (_,e,_) in islice(zstream,401)]
plt.plot(dt * np.array(range(len(elist))), elist)
ax.set_ylabel(r"Error term, $e(t)$", fontsize=18)
ax.set_xlabel(r"Time, $t$", fontsize=18)
plt.show()
\end{python}
\noindent
which plots the evolution of the error term as depicted in \figref{fig:diagram}.

%\newpage 

Readers familiar with the Fourier transform may recognize that if the input features $x_k$  are sines and cosines as in the example above, the adaptive combiner computes the Fourier coefficients for $y$ in terms of $\myvec{x}$.

We can run the next iteration of the adaptive linear combiner to check the weights
by executing \pyth{next(zstream)}, which returns
\begin{python}[basicstyle=\scriptsize\ttfamily]
(0.039, 0.001,
 array([ 0.810,  0.   , -0.090,  0.   ,  0.033,  0.   ,  -0.017,  0.   ,  
         0.010,  0.   , -0.007,  0.   ,  0.005,  0.   ,  -0.004,  0.   ,
         0.003,  0.001, -0.002]))
\end{python}

The array contains the weights $w_k$. They agree well with the theoretically optimal values, which are the trigonometric Fourier coefficients for the triangle wave, and which can be computed as follows:

\begin{python}[basicstyle=\scriptsize\ttfamily]
np.array([(8/np.pi**2/x**2*(-1)**((x-1)/2) if x%2==1 else 0) 
for x in range(1,20)])	

array([ 0.811,  0.   , -0.09 ,  0.   ,  0.032,  0.   , -0.017,  0.   ,
    0.01 ,  0.   , -0.007,  0.   ,  0.005,  0.   , -0.004,  0.   ,
    0.003,  0.   , -0.002])
\end{python}

However, the adaptive linear combiner is more general than the Fourier transform, because the feature vector components $x_k(t)$ are not required to be orthogonal.
Non-orthogonal component signals $x_k(t)$ can still be used, but convergence may be slower. 

\subsubsection{* Theory}
As noted at the beginning of \sectref{sec:basic-approx}, the fundamental function of the adaptive linear combiner is to use a set of provided basis functions $x_1(t), x_2(t),\ldots x_n(t)$ to approximate a function $y(t)$,
\begin{equation}
    y(t) \approx \sum_{k=1}^n w_k x_k ,
%    w_1 x_1(t) + w_2 x_2(t) +\ldots + w_k x_k(t)  +\ldots + w_n x_n(t) ,
\end{equation}
where $w_1, w_2,\ldots w_n$ are the weights to be found by the adaptive combiner.
Formally, the weights should minimize the error (residual) norm
\begin{equation}
\label{eq:err:norm}
    \left\| y - \sum_{k=1}^n w_k x_k\right\|^2 ,
\end{equation}
where the norm $\|f\|$ of a function $f(t)$ is defined by
\begin{equation}
    \left\|f\right\|^2 = \int{f^2(t) dt} ,
\end{equation}
where the integral is taken over the interval of interest.

A simple example is when the basis functions $x_k(t)$ are the trigonometric functions
$1, \cos t, \sin t, \cos 2 t, \sin 2 t,\ldots$ as in the example in \sectref{sec:basic-approx} above. In this case, the optimal weights $w_k$ are the trigonometric Fourier coefficients of $y$.

The basis functions $x_k(t)$ can be almost any kind of function, but the adaptive combiner will work better if they are ``similar'' (in a mathematical sense) to the reference signal $y(t)$. 
For the efficiency of the adaptive combiner, it is essential to choose suitable basis functions. 

For the optimal weights to be unique, the basis functions must be linearly independent, a condition that can be difficult to satisfy in practice. In the case of an adaptive combiner, however, uniqueness of the weights is generally not required. What matters is that the error is minimized.

When deployed, the adaptive combiner will only be given data for discrete time points
$t=0, 1,2,\ldots$. If $\myvec{x}(t)$ and $y(t)$ samples are available simultaneously for all these times, the problem of minimizing the error norm \eqref{eq:err:norm} turns into the standard least squares optimization problem. Suppose that values of the reference signal are provided as a vector $\myvec{y} = \left(y(0), y(1),\ldots, y(t)\right)^\top$ and that the activations of the basis functions are given as the matrix
\begin{equation}
 \myvec{X} = \begin{bmatrix}
 \myvec{x}(0)^\top\\
 \myvec{x}(1)^\top\\
 \vdots\\
 \myvec{x}(t)^\top\\
\end{bmatrix} .
\end{equation}
Then the optimal weights $\myvec{w}_{\texttt{opt}}$ can be found from the overdetermined equation system
\begin{equation}
    \myvec{X} \myvec{w} = \myvec{y}.
\end{equation}
A least squares solution to this system can be computed as:
\begin{equation}
    \myvec{w}_{\texttt{opt}} = \myvec{X}^+ \myvec{y},
    \label{eq:w:pseudo}
\end{equation}
where $\myvec{X}^+$ is the pseudoinverse of $\myvec{X}$, which can be found by singular-value decomposition.
However, if the adaptive combiner receives only one row of $\myvec{X}$ at a time, it can update the weights incrementally each time $\myvec{x}(t)$ arrives (cf. \algref{alg:adaptive-linear-combiner}). As noted above, this process corresponds to the stochastic gradient descent method, and will eventually converge to the same error norm as the ``batch'' solution in \eqref{eq:w:pseudo}.

%\newpage

\begin{figure}[!t]
	\centering{\includegraphics[width=0.7 \textwidth]{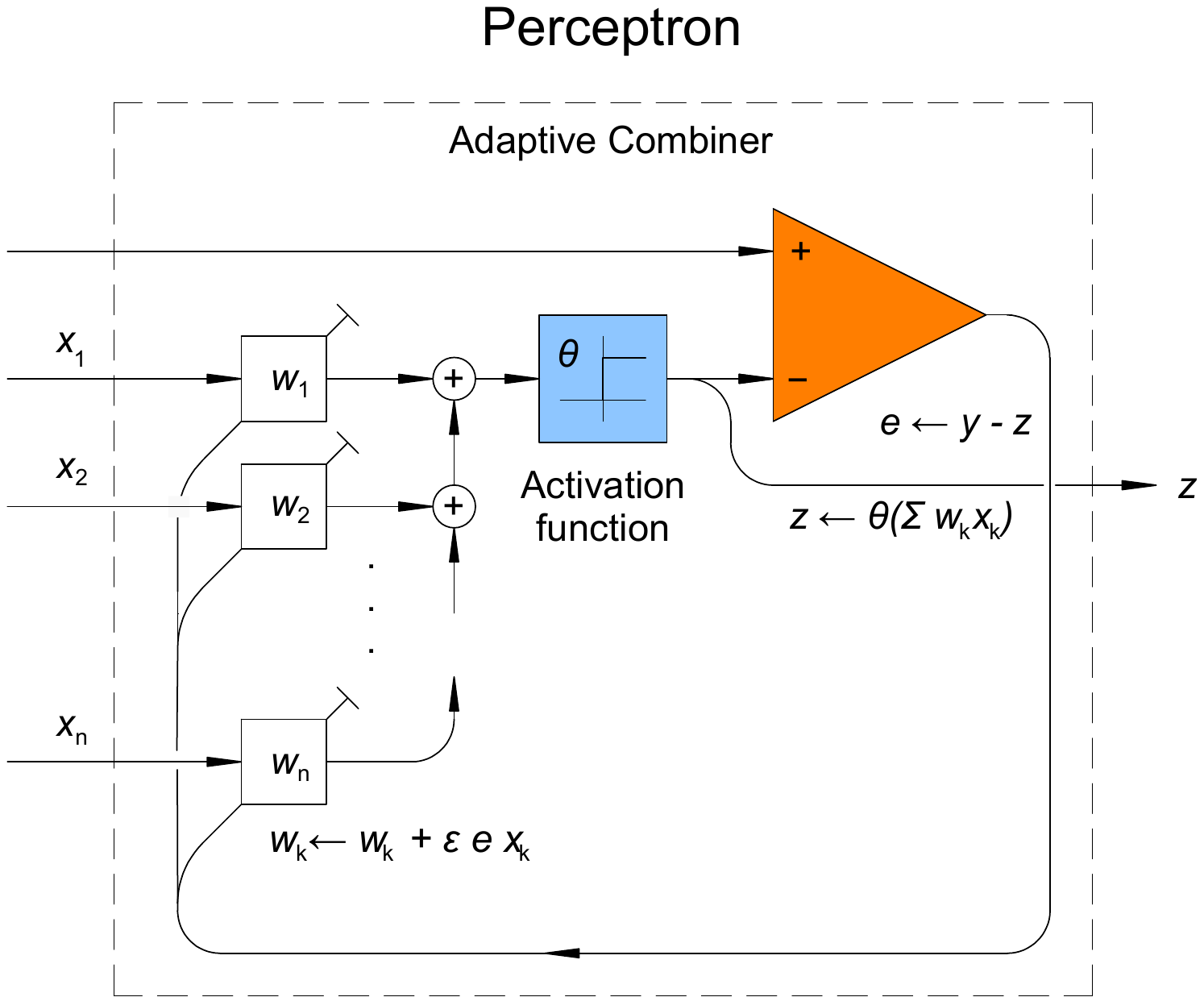}}
	\caption{{\bf The Perceptron.} The Perceptron was one of the very first neural networks. The output from the Perceptron is binary, i.e., either zero or one. The dashed box outlines the extent of the adaptive combiner. The blue shaded area highlights the main difference (the activation function) from the adaptive linear combiner in \figref{fig:adaptive-linear-combiner}.  
    }
    \label{fig:perceptron}
\end{figure}

\subsection{The Perceptron}

The Perceptron (\figref{fig:perceptron}) was one of the very first neural networks, and predates the adaptive linear combiner. Its neuron model was proposed in 1943 by McCulloch and Pitts~\citep{McCulloch.Pitts.1943alc} while Rosenblatt~\citep{Rosenblatt.1958tpa} developed the ideas of learning in networks of such neurons and implemented these ideas in hardware.

The principal difference from the adaptive linear combiner is that the prediction $z(t) = \myvec{w}^\top \myvec{x}(t)$ is fed through an {\em activation function} before being output and used to calculate the error  (see \figref{fig:perceptron}). In the case of the Perceptron, the activation function is Heaviside's step function $\theta(z(t))$, which outputs zero if the argument $z(t)$ is negative and one otherwise.
Pseudocode for the Perceptron is shown in \algref{alg:perceptron}.

\begin{algorithm}[!]
	\caption{{\bf The Perceptron}. The Perceptron requires a bias term, e.g., $x_1(t)$, to be fixed to one: $x_1(t) = 1$. As an activation function $\theta(\cdot)$, it uses Heaviside's step function. Otherwise, it is identical to the adaptive linear combiner, \algref{alg:adaptive-linear-combiner}.}
    \label{alg:perceptron}
	\SetKwInOut{Input}{input}
	\SetKwInOut{Constant}{constant}
	\SetKwInOut{Local}{local}
	\SetKwInOut{Output}{output}
	\SetKwComment{Comment}{/* }{ */}
	\Input{$x_k(t), 1\le k\le n$ \Comment*[r]{Component signals} }
	\Input{$y(t)$ \Comment*[r]{Reference signal}}
	\Constant{$n \ge 1$ \Comment*[r]{Number of components}}
	\Constant{$\varepsilon > 0$ \Comment*[r]{Learning rate}}
	\Output{$z(t)$  \Comment*[r]{Prediction}}
	\Output{$e(t)$  \Comment*[r]{Error}}
	\Local{$w_k, 1\le k\le n$  \Comment*[r]{Weights} }
	\For{$t \gets 0, 1,\ldots$}{   
		$z(t) \gets \theta\left[\Sigma_{k=1}^n w_k x_k(t)\right]$ \Comment*[r]{Compute prediction}
		$e(t) \gets y(t) - z(t)$ \Comment*[r]{Compute error}
		\For {$k \gets 1,\ldots, n$} {
			$w_k \gets w_k + \varepsilon \, e(t) x_k(t)$ \Comment*[r]{Update weights}
		}
	}
\end{algorithm} 

%\clearpage
\newpage
\subsection{ADALINE}

\begin{figure}[!ht]
	\centering{\includegraphics[width=0.7 \textwidth]{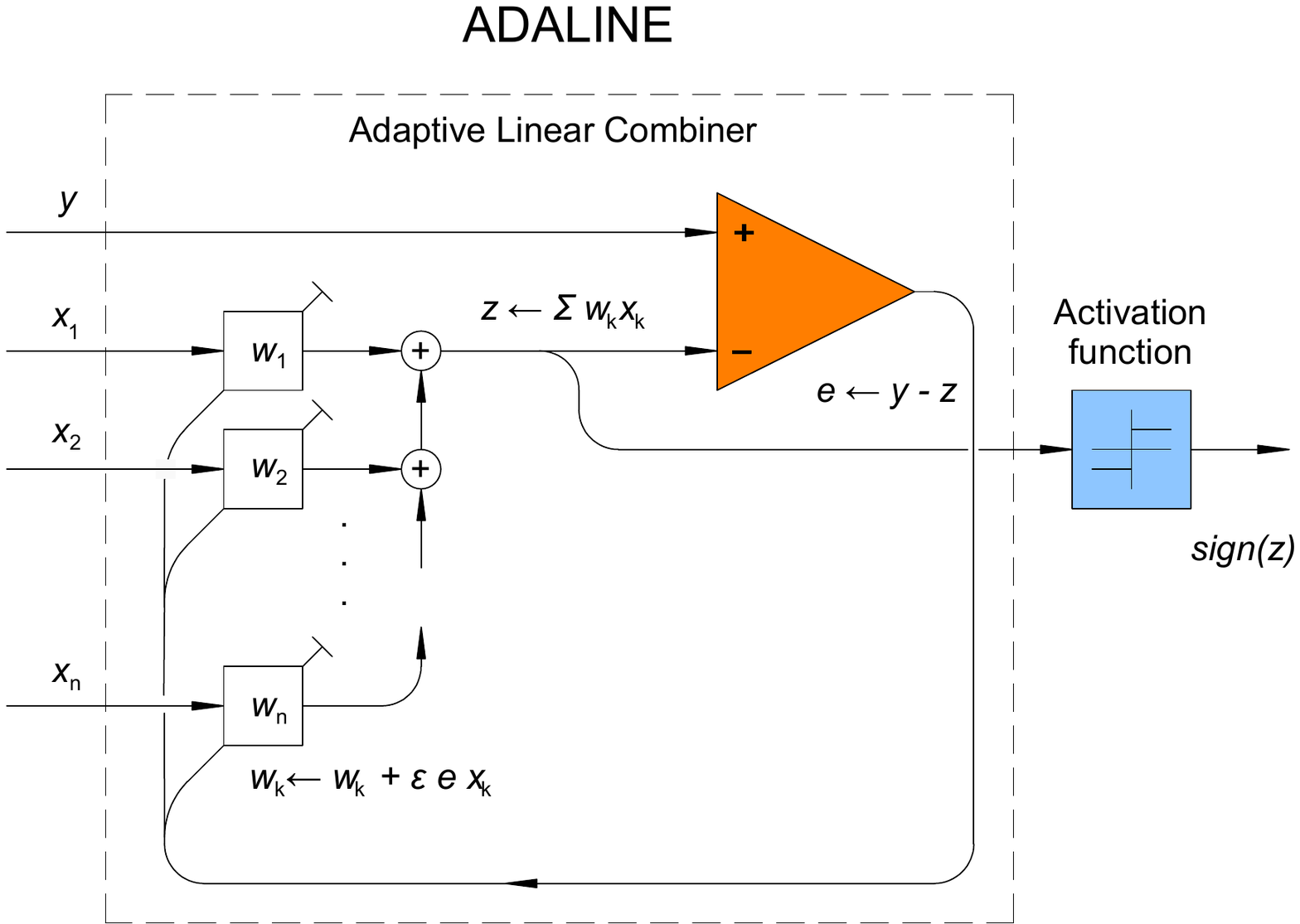}}
	\caption{{\bf ADALINE.} Originally a neuron model, ADALINE traces back to 1960. %Again, 
    The dashed box outlines the extent of the adaptive combiner, and the blue shaded area highlights the main difference from the perceptron in \figref{fig:perceptron}.
		\label{fig:adaline}}
\end{figure}

ADALINE illustrated in 
\figref{fig:adaline} was originally an acronym for ``ADAptive LInear NEuron'', and was proposed as a neuron model by
Widrow and Hoff~\citep{Widrow.Hoff.1960asc}. It is similar to the Perceptron, but the error term is computed
before applying the activation function, and the sign function is used as an activation function instead of Heaviside's step function. 

Pseudocode for ADALINE is listed as \algref{alg:adaline}. 
For a more detailed comparison of ADALINE and the Perceptron, please refer to~\citep{Widrow.Lehr.19903yo}. 
This study also provides some interesting history of the relation between ADALINE and backpropagation.

\begin{algorithm}[!ht]
	\caption{{\bf ADALINE}. The difference between the Perceptron and ADALINE is that in ADALINE, the error term is computed {\em before} $\sign(\cdot)$ activation function is applied.}
    \label{alg:adaline}
	\SetKwInOut{Input}{input}
	\SetKwInOut{Constant}{constant}
	\SetKwInOut{Local}{local}
	\SetKwInOut{Output}{output}
	\SetKwComment{Comment}{/* }{ */}
	\Input{$x_k(t), 1\le k\le n$ \Comment*[r]{Component signals} }
	\Input{$y(t)$ \Comment*[r]{Reference signal}}
	\Constant{$n \ge 1$ \Comment*[r]{Number of components}}
	\Constant{$\varepsilon > 0$ \Comment*[r]{Learning rate}}
	\Output{$z(t)$  \Comment*[r]{Prediction}}
	\Local{$e(t)$  \Comment*[r]{Error}}
	\Local{$w_k, 1\le k\le n$  \Comment*[r]{Weights} }
	\For{$t \gets 0, 1,\ldots$}{   
		$z(t) \gets \sign\left[\Sigma_{k=1}^n w_k x_k(t)\right]$ \Comment*[r]{Compute prediction}
		$e(t) \gets y(t) -\Sigma_{k=1}^n w_k x_k(t)$ \Comment*[r]{Compute error}
		\For {$k \gets 1,\ldots, n$} {
			$w_k \gets w_k + \varepsilon \, e(t) x_k(t)$ \Comment*[r]{Update weights}
		}
	}
\end{algorithm}

%\newpage
%\phantom{a}
%\newpage

\subsection{The adaptive filter}
\label{sec:adaptive:filter}

\begin{figure}[!ht]
	\centering{\includegraphics[width=0.7 \textwidth]{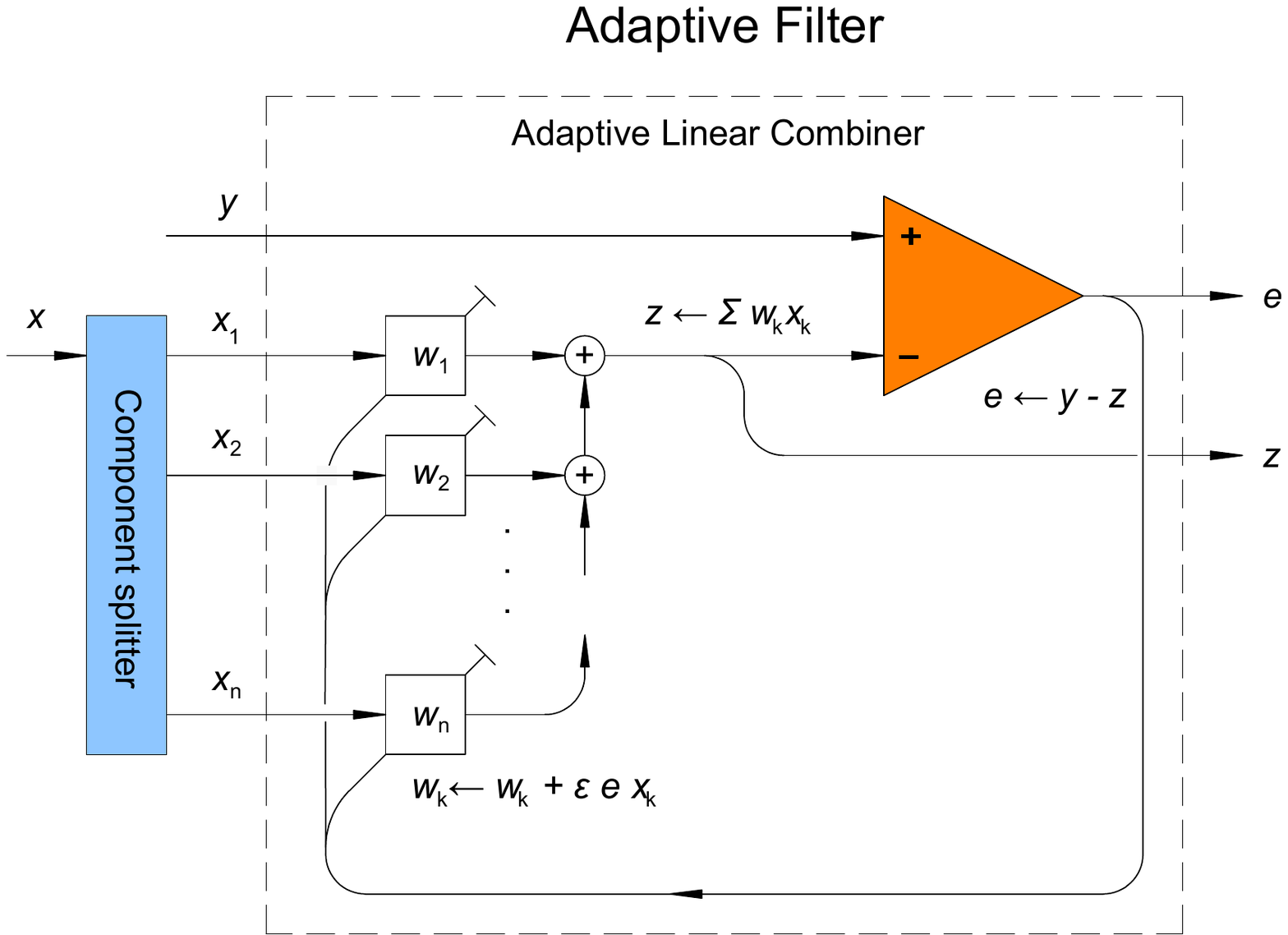}}
	\caption{{\bf The adaptive filter.}
		\label{fig:adaptive-filter} The adaptive filter is an adaptive linear combiner with a component splitter. The component splitter
        splits the input signal $x$ into ideally orthogonal, or at least linearly independent, component signals $x_k$. The dashed box outlines the adaptive combiner extents, and the blue shaded area highlights the main difference (the component splitter) from the adaptive linear combiner in \figref{fig:adaptive-linear-combiner}.
        }
\end{figure}

Widrow and Hoff eventually abandoned the idea of modeling the neuron as efforts to develop learning rules for networks with multiple adaptive layers were unsuccessful. 
They, however, reinterpreted ADALINE as ``ADAptive LINear Element'', and the adaptive linear combiner turned out to be tremendously useful in technical applications. 
It became the start of a whole new research field, now known as adaptive signal processing~\citep{Haykin.2002aft,Widrow.Stearns.1985asp}, which provides powerful building blocks for signal processing algorithms. 
Depending on how the adaptive filter is connected, it can generate a model or an inverse model for a given signal, predict the future of a signal, or cancel noise from a signal~\cite{Widrow.Stearns.1985asp}.

As the adaptive linear combiner, the adaptive filter (illustrated in \figref{fig:adaptive-filter}) does not apply any activation function to $\myvec{z}(t)$. 
However, it differs from the adaptive linear combiner by having a scalar $x(t)$ input instead of a vector $\myvec{x}(t)$.
Instead, the single $x(t)$ is split into components $x_k(t)$, {\it e.g.}, by a tapped delay line, a bandpass filter bank, or a running Fourier transform.
Ideally, the split-out component signals should be orthogonal, leading to faster convergence, but the adaptive filter will operate properly even if they are non-orthogonal. 
\algref{alg:adaptive-filter} presents the pseudocode for the adaptive filter with the tapped delay line.

\begin{algorithm}[!ht]
	\caption{{\bf Adaptive filter}. The adaptive filter is an adaptive linear combiner 
		where the component signals $x_k(t)$ are extracted from a scalar input stream $x(t)$.
		In this algorithm, they are generated via a delay line.}\label{alg:adaptive-filter}
	\SetKwInOut{Input}{input}
	\SetKwInOut{Constant}{constant}
	\SetKwInOut{Local}{local}
	\SetKwInOut{Output}{output}
	\SetKwComment{Comment}{/* }{ */}
	\Input{$x(t)$ \Comment*[r]{Input signal from which components are generated} }
	\Local{$x_k(t), 1\le k\le n$ \Comment*[r]{Component signals} }
	\Input{$y(t)$ \Comment*[r]{Reference signal}}
	\Constant{$n \ge 1$ \Comment*[r]{Number of components}}
	\Constant{$\varepsilon > 0$ \Comment*[r]{Learning rate}}
	\Output{$z(t)$  \Comment*[r]{Prediction}}
	\Output{$e(t)$  \Comment*[r]{Error}}
	\Local{$w_k, 1\le k\le n$  \Comment*[r]{Weights} }
	\For{$t \gets 0, 1,\ldots$}{
		$x_1(t) \gets x(t)$ \Comment*[r]{Update to new value}
		\For{$k \gets 2,\ldots, n$}{
			$x_k(t) \gets x_{k-1}(t-1)$ \Comment*[r]{Update to delayed values}
		}   
		$z(t) \gets \Sigma_{k=1}^n w_k x_k(t)$ \Comment*[r]{Compute prediction}
		$e(t) \gets y(t) - z(t)$ \Comment*[r]{Compute error}
		\For {$k \gets 1,\ldots, n$} {
			$w_k \gets w_k + \varepsilon \, e(t) x_k(t)$ \Comment*[r]{Update weights}
		}
	}
\end{algorithm}

\newpage
%\phantom{a}
%\newpage

\subsection{The adaptive conical combiner}

\begin{figure}[!ht]
	\centering{\includegraphics[width=0.91 \textwidth]{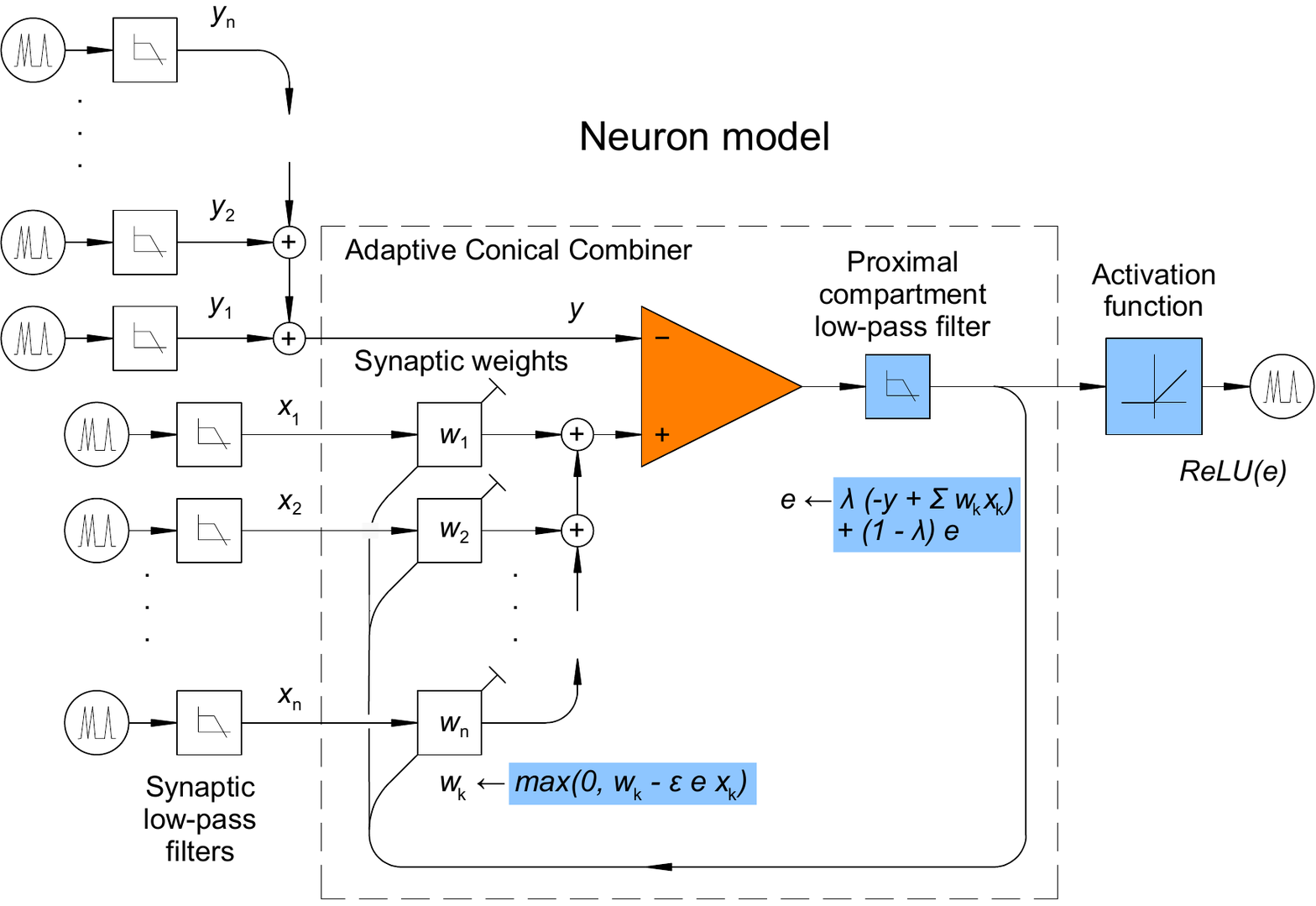}}
	\caption{{\bf The adaptive conical combiner.} Here, the adaptive conical combiner is shown as part of a neuron model. The model is biologically accurate and derived directly from how biological neurons operate. The principal difference from all previous adaptive combiner variants is that weights are non-negative. Note also that the sign of the error is opposite to that of the previous combiners. This is because the reference signal for biological neurons is the inhibitory (inverting) input. The encircled curves symbolize neuronal spike trains.
The dashed box outlines the adaptive combiner extents, and the blue shaded areas highlight the main differences from the adaptive filter in \figref{fig:adaptive-filter} (adapted from
{\protect{\cite{Nilsson.2023ipb}}}).
		\label{fig:adaptive-conical-combiner}}
\end{figure}

The adaptive conical combiner (\figref{fig:adaptive-conical-combiner}), based on the biologically accurate neuron model described in~\citep{Nilsson.2023meo}, differs from the previous adaptive combiners mainly by enforcing non-negative weights to reflect the properties of biological synapses. 
This requires a weight update rule that includes a check to prevent the weights from becoming negative. 
In addition, it incorporates a low-pass filter on the error before feedback.
Pseudocode for the adaptive filter is shown in \algref{alg:adaptive-conical-combiner}.

When the adaptive conical combiner is used in a neuron model, the reference signal $y$ corresponds to the summed inhibitory inputs, and the $x_k$ inputs to the excitatory inputs. The only output is the filtered error term $e$. It may seem as a restriction that there is no prediction term $z$, but this signal can be recreated by combining together two adaptive conical combiners~\citep{Nilsson.2023ipb}.

The non-negativity of weights carries a special significance. While at first this non-negativity seems to be a restriction, it allows the representation and processing of sparse hierarchical data structures in a more efficient way than signed weights~\citep{Nilsson.2023ipb}.

\begin{algorithm}[!t]
	\caption{{\bf Adaptive conical combiner}. The main difference from the adaptive linear combiner is that
	weights cannot become negative. In addition, the error is low-pass filtered
    (from {\protect{\cite{Nilsson.2023meo}}}).
    }
    \label{alg:adaptive-conical-combiner}
	\SetKwInOut{Input}{input}
	\SetKwInOut{Constant}{constant}
	\SetKwInOut{Local}{local}
	\SetKwInOut{Output}{output}
	\SetKwComment{Comment}{/* }{ */}
	\Input{$x_k(t), 1\le k\le n$ \Comment*[r]{Component signals} }
	\Input{$y(t)$ \Comment*[r]{Reference signal}}
	\Constant{$n \ge 1$ \Comment*[r]{Number of components}}
	\Constant{$\varepsilon > 0$ \Comment*[r]{Learning rate}}
	\Constant{$\lambda \in \, ]0,1]$ \Comment*[r]{Error filter parameter}}
	\Output{$z(t)$  \Comment*[r]{Prediction}}
	\Output{$e(t)$  \Comment*[r]{Error}}
	\Local{$w_k, 1\le k\le n$  \Comment*[r]{Weights} }
	\For{$t \gets 0, 1,\ldots$}{   
		$z(t) \gets \Sigma_{k=1}^n w_k x_k(t)$ \Comment*[r]{Compute prediction}
		$e(t) \gets \lambda \left[- y(t) + z(t)\right] + (1 - \lambda) e(t-1)$ \Comment*[r]{and error}
		\For {$k \gets 1,\ldots, n$} {
			$w_k \gets \max\left[0, w_k - \varepsilon \, e(t) x_k(t)\right]$ \Comment*[r]{Update weights}
		}
	}
\end{algorithm}

\clearpage
\section{Cascaded adaptive combiners}
\label{sec:cacs}
\begin{figure}[!ht]
	\centering
    \includegraphics[width=1.00\linewidth]{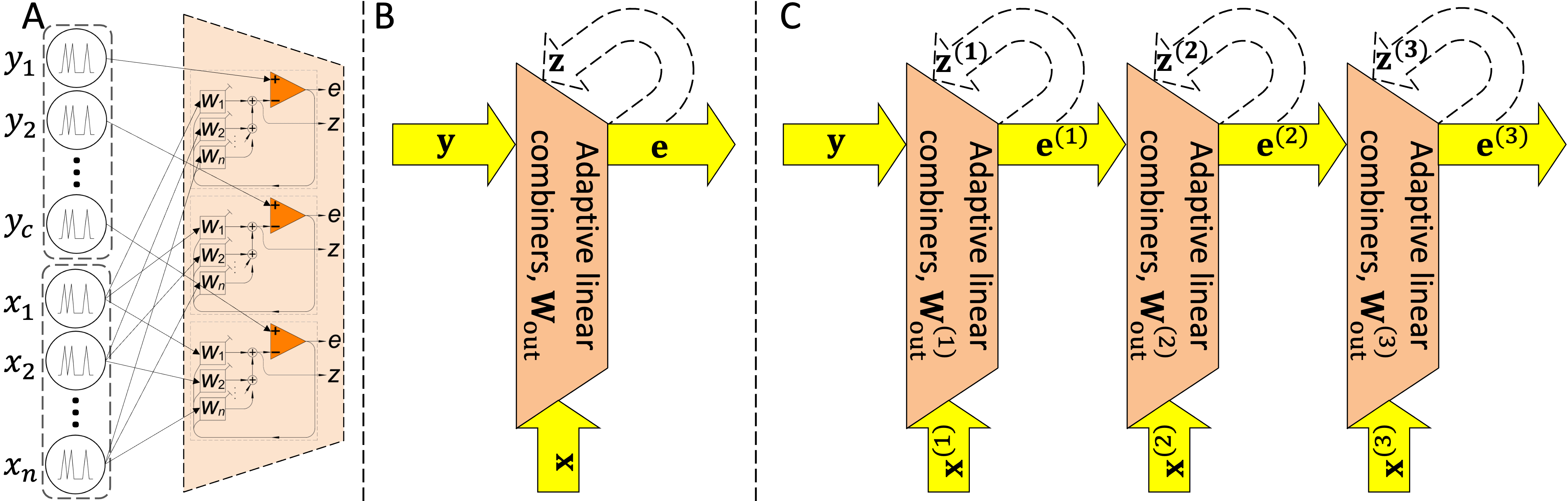}
	\caption{{\bf Cascaded adaptive combiners.} 
		{\bf A.} A layer is formed by multiple adaptive combiners operating in parallel.
		The inputs and outputs $\myvec{y}$, $\myvec{x}$, and $\myvec{e}$ are vectors.
		{\bf B.} We use a trapezoid as a symbol for the layer in panel A.
		{\bf C.} A three-layer CAC: the error term from one layer provides the reference signal to the next layer. CACs perform successive approximations, trying to eliminate the error term using different transformations of the input signals (${\bm x}^{(1)}$, ${\bm x}^{(2)}$, ${\bm x}^{(3)}$) that are computed from, e.g., different basis functions (not shown in this figure). 
	}
	\label{fig:cac}
\end{figure}

By stacking layers of adaptive combiners, we create networks capable of learning, CACs. 
In this section, we describe the individual layers, the cascading of these layers, and the learning process in more detail.

\subsection{Single layers of adaptive combiners}

Adaptive combiners can be organized into layers (\figref{fig:cac}A), analogous to those used in conventional neural networks. 
The key distinction is that we retain the differential nature of the system by maintaining  separate inhibitory and excitatory inputs.
Inhibitory inputs are used to provide reference signals, while excitatory inputs carry the feature vectors.
We symbolically represent such a layer as illustrated in \figref{fig:cac}B; both the inputs and outputs are vectors.

\subsubsection{Programming}
\label{sec:CAC:layer:progr}

The implementation for a single adaptive linear combiner (\sectref{sec:basic-approx}) can be extended to construct a layer of multiple adaptive combiners. 
This is achieved by computing the weights using the least-squares solution (cf. \eqref{eq:cac:ls} in \sectref{sec:CAC:layer:theory}):

\begin{python}[basicstyle=\scriptsize\ttfamily]
import numpy as np

def train_adaptive_combiners(X, Y = None):
    """ Adapt combination of X-components to Y and
	    generate the output streams. """
    # Here using pseudo-inverse (convenient for small-dim X)
    # Could also be done incrementally by gradient descent
    W = np.linalg.pinv(X) @ Y
    # Output error/residual and weights
    return (Y - X @ W, W)
    
def test_adaptive_combiners(X, W = None):
    """ Test adaptive combiners by clamping Y to zero. """
    # Assume Y = 0 and output error/residual (= -prediction)
    return (- X @ W)
\end{python}

\subsubsection{* Theory}
\label{sec:CAC:layer:theory}

The case of a single adaptive combiner can easily be generalized to a layer of multiple parallel adaptive combiners. Here, the reference signal becomes a matrix $\myvec{Y}$, having one row for every time point and one column for each adaptive combiner's reference signal. The expression to be minimized is
\begin{equation}
    \left\| \myvec{Y} - \myvec{X} \myvec{W} \right\|^2 ,
\end{equation}
where $\myvec{W}$ is a weight matrix, with one column for each adaptive combiner in the layer and one row for each basis function. This system has a least squares solution
\begin{equation}
    \myvec{W}_{\texttt{opt}} = \myvec{X}^+\myvec{Y}.
    \label{eq:cac:ls}
\end{equation}
The adaptive combiners can also solve this system incrementally by running stochastic gradient descent in parallel for all combiners in the layer. Such a solution is useful when $\myvec{X}$ is very large or a stream of input data. A full code example 
can be found in the appendix.

After training a layer, we want to test it, {\it i.e.}, we want to see its prediction given a row vector $\myvec{x}$ of basis function data. This can be performed by feeding the reference signal $\myvec{y} = 0$ and disabling weight updates by setting the learning rate to zero. The output of the layer will be the row vector
\begin{equation}
    \myvec{e} = \myvec{y} - \myvec{x} \myvec{W}   = - \myvec{x} \myvec{W},
    \label{eq:cac:resid}
\end{equation}
which is the prediction with the opposite sign.

\subsection{Cascaded layers of adaptive combiners}

CAC are constructed by stacking layers of adaptive combiners, \figref{fig:cac}C, with multiple possible configurations. While the most general form of CAC uses layers of adaptive conical combiners with non-linear activation functions, it is often practical to use layers of simple adaptive linear combiners. Here, we demonstrate CAC with adaptive linear combiners.

Although no activation function is applied between layers of adaptive combiners, CAC nevertheless exhibits non-linear behavior through transformations of the input signal. Such non-linearity is indispensable in any non-trivial CAC architecture; in its absence, the network would be equivalent to a single-layer neural network.

A closely related central issue for CAC concerns the selection of basis functions. In a single-layer setting, an arbitrary random transformation may be sufficient. In deeper architectures, however, each successive layer must introduce a new set of basis functions that is not merely a linear transformation of those in the preceding layer, since such a transformation would not enable the removal of components that could not already be eliminated earlier. Accordingly, the basis functions in each new layer must arise through a non-linear transformation of those from the previous layer.

A simple means of achieving this is to multiply the basis functions from the preceding layer by a random matrix and subsequently apply a ReLU non-linearity. This approach was adopted in the programming examples presented in this article and yielded good performance on the MNIST benchmark. More generally, however, the choice of basis functions is a critical determinant of CAC performance.
 
\subsubsection{Programming}
\figref{fig:cac}C shows three layers of adaptive combiners, which we assume to be linear. For the first layer, the reference signal is $\myvec{y}$ and the feature vector is $\myvec{x}^{(1)}$. The output of this layer is the error term, or residual, remaining after the reference signal has been approximated by the weight matrix of the first layer,
\begin{equation}
\label{eq:first-residual}
    \myvec{e}^{(1)} = \myvec{y} - \myvec{x}^{(1)} \myvec{W}^{(1)} 
\end{equation}
%, where $\myvec{W}^{(1)}$ is the weight matrix.

In the second layer, the vector of error terms from the first layer becomes the new reference signal $\myvec{y}^{(2)}=\myvec{e}^{(1)}$,
\begin{equation}
\label{eq:second-residual}
    \myvec{e}^{(2)} = \myvec{y}^{(2)} - \myvec{x}^{(2)} \myvec{W}^{(2)} = \myvec{e}^{(1)} - \myvec{x}^{(2)} \myvec{W}^{(2)} = \myvec{y} - \myvec{x}^{(1)} \myvec{W}^{(1)} -  \myvec{x}^{(2)} \myvec{W}^{(2)}
\end{equation}
Crucially, the new feature vector $\myvec{x}^{(2)}$ must be generated from the original input signal $\myvec{x}$. This transformation must be non-linear, as a linear transformation would not be able to reduce the error term further than what was achieved by the previous layer. We can think of $\myvec{x}^{(k)}$ as providing a basis for the error term $\myvec{e}^{(k-1)}$, and whatever cannot be accounted for by this basis becomes a new error term, $\myvec{e}^{(k)}$. From $\myvec{x}$, CAC generates a new basis $\myvec{x}^{(k+1)}$, which can eliminate most of the new error term, and this process continues.

Importantly, the convergence of the weights $\myvec{W}^{(1)}$ in the first layer does not depend on the convergence of any subsequent layers. 
The same property holds for every layer, ensuring that the overall training procedure converges.
To predict the reference signal $\myvec{y}$ from a given $\myvec{x}$, we simply set $\myvec{y} = 0$ and compute the sum of the error terms as defined in  \eqref{eq:cac:resid}. 
This sum corresponds to the prediction term with the opposite sign, $-\myvec{z}$.

The least-squares solution in \eqref{eq:cac:ls} assumes a batch formulation, where the weights are computed from the entire dataset at once. 
In many practical settings, however, data arrive sequentially and the CAC must adapt incrementally. 
In such cases, the weights can be updated online using either the LMS rule above or recursive least-squares, which provides an efficient incremental alternative that converges to the same least-squares optimum while updating the parameters as new samples become available. This allows adaptive combiner layers to operate in streaming scenarios while retaining the convergence guarantees of the batch formulation.

Using the primitive for a single layer of adaptive combiners (\sectref{sec:CAC:layer:progr}), we can easily chain layers together to form a CAC, assuming the batch formulation:
\begin{python}[basicstyle=\scriptsize\ttfamily]
def cascaded_adaptive_combiners(X, Y = None, W = None, train = True, 
                                layers = 10, expansion_size = 1296, seed = 42):
    """
    Cascaded Adaptive Combiners (CAC).

    Two modes:
    - Train (train = True): Weights are learned for each layer.
    - Test (train = False): Prediction is computed using given weights.
    """
    
    # Generate random projection matrix
    rand_proj = np.random.default_rng(seed).standard_normal((expansion_size, 
                                                             expansion_size))
    if train: 
        # Initialize error/resídual and weights for training mode
        E = Y
        W = []
    else: # test
        # Initialize prediction to zero for test mode
        prediction = 0
    
    print("\nCurrent layer:", end="")
    for layer in range(layers):
        print(f" {layer}", end="", flush = True)
        # Compute new basis for current layer by random projection
        # of previous layer basis and ReLU nonlinearity
        X = np.maximum(0, X@rand_proj[:X.shape[1],:]) 

        if train:        
            # Compute weight matrix for current layer
            E, W_current = train_adaptive_combiners(X, Y = E)
            W.append(W_current)
        else: # test
            # Compute error/residual after current layer
            E = test_adaptive_combiners(X, W = W[layer])            
            # Update prediction
            prediction -= E

    # Output weigths or prediction depending on mode
    return (W if train else prediction)
\end{python}
The following is an example which first loads the MNIST benchmark \cite{Deng.2012tmd} and then applies
the cascaded adaptive combiner to compute the accuracies for the training set and test set, respectively:
\begin{python}[basicstyle=\scriptsize\ttfamily]
from torchvision import datasets

# Load and scale training data, one-hot encoding targets
trainset = datasets.MNIST(root = './data', train = True, download = True)
X_train = trainset.data.float().div(255).view(len(trainset), -1).numpy()
Y_train = trainset.targets.numpy()
Y_train_onehot = np.eye(10, dtype=np.float32)[Y_train]

# Load and scale test data
testset = datasets.MNIST(root = './data', train = False, download = True)
X_test = testset.data.float().div(255).view(len(testset), -1).numpy()
Y_test = testset.targets.numpy()

# Train weights 
weights = cascaded_adaptive_combiners(X_train, Y = Y_train_onehot, train = True)

# Measure accuracy
for (mode, X, Y) in [("Train", X_train, Y_train), ("Test", X_test, Y_test)]:
    prediction = cascaded_adaptive_combiners(X, W = weights, train = False)
    accuracy = np.mean(np.argmax(prediction, axis = 1) == Y)
    print(f"\n{mode} accuracy: {accuracy:.4f}")
\end{python}

The output from the program is as follows:

\begin{python}[basicstyle=\scriptsize\ttfamily]
Current layer: 0 1 2 3 4 5 6 7 8 9
Current layer: 0 1 2 3 4 5 6 7 8 9
Train accuracy: 0.9823

Current layer: 0 1 2 3 4 5 6 7 8 9
Test accuracy: 0.9662
\end{python}

\subsubsection{* Theory}

As we have seen above, the first layer computes the error term $\myvec{e}^{(1)}$.
The second layer treats this error term as the reference signal instead of $\myvec{y} $, and uses a new set of features $\myvec{x}^{(2)}$ to approximate the given error term.
The third layer will then try to approximate the new error term $\myvec{e}^{(2)}$ using $\myvec{x}^{(3)}$, and so on, successively reducing the error terms towards zero. In other words, extending identities \eqref{eq:first-residual}-\eqref{eq:second-residual} above,
\begin{align}
    \myvec{y} &= \myvec{e}^{(1)} + \myvec{x}^{(1)} \myvec{W}^{(1)} \notag \\
    &= \left[\myvec{e}^{(2)} + \myvec{x}^{(2)} \myvec{W}^{(2)} \right]
       + \myvec{x}^{(1)} \myvec{W}^{(1)}\notag \\
    &= \left[\myvec{e}^{(3)} + \myvec{x}^{(3)} \myvec{W}^{(3)} \right]
       + \myvec{x}^{(2)} \myvec{W}^{(2)}
       + \myvec{x}^{(1)} \myvec{W}^{(1)}\notag \\
       &\ldots \notag \\
    &= \sum_{k=1}^\infty \myvec{x}^{(k)} \myvec{W}^{(k)} . 
    \label{eq:prediction}
\end{align}
Therefore, in the prediction phase, we can approximate this expression by setting all reference signals to zero, disabling weight updates, and then forming the negative of the total of the error terms.

\begin{figure}[!ht]
	\centering
    \includegraphics[width=0.95 \linewidth]{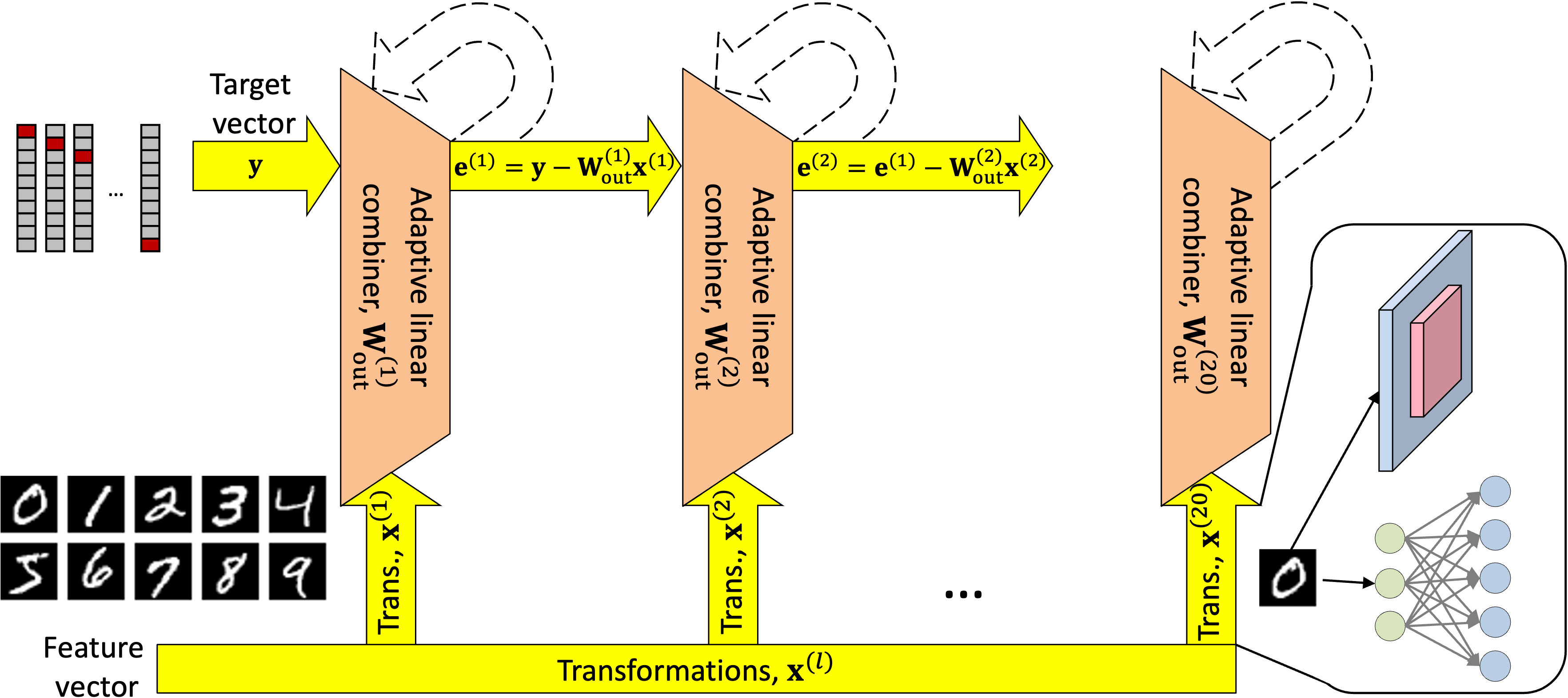}
	\caption{{\bf Cascaded adaptive combiners for classifying MNIST images}. 
	The network follows the structure in~\figref{fig:cac}C, stacking up to $20$ layers of adaptive linear combiners. Shown $28 \times 28$ pixels images are examples of MNIST digits used as the input to CAC. The  reference signal is a one-hot encoding corresponding to the digit in the input image.  
    In the experiments, we evaluated two ways of transforming MNIST images into feature vectors: random projects followed by $\mathrm{ReLU}$ activation function and random convolutional filters followed by max pooling.
	}
\label{fig:mnist:cac}
\end{figure}

\begin{figure}[!ht]
	\centering
    \includegraphics[width=0.8 \linewidth]{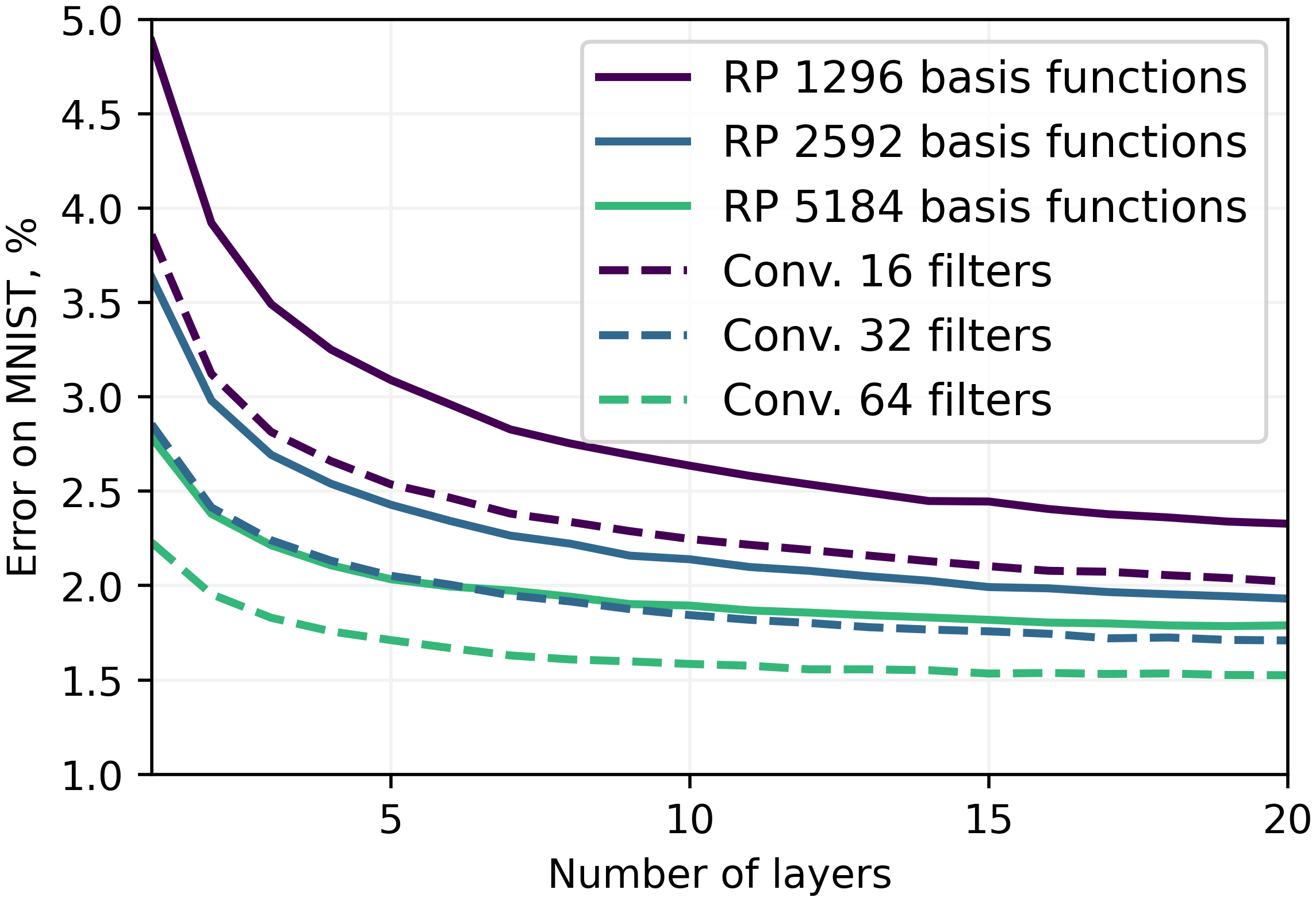}
	\caption{{\bf Prediction of CACs on MNIST}. The figure illustrates the prediction error on MNIST testing images against different number of layers in CACs. 
    Colors denote different number of features per layer while dash types indicate two different transformations used in the experiments: random projections (RP in the legend) and random convolutions (Conv. in the legend).       
    %For all layers the filter size was fixed to $10 \times 10$ pixels. 
    The reported results are averaged over $30$ simulation runs. 
	}
	\label{fig:mnist:error}
\end{figure}

\section{Digit classification: The MNIST example}
\label{sec:combiners:mnist}

To illustrate building machine learning models with CACs and the effect of their design choices on the overall performance of the model, we applied CAC to the MNIST dataset -- a well-known benchmark of classifying hand-written digits based on their $28 \times 28$ pixels images. 
The explored models are constructed using adaptive linear combiners (\sectref{sec:linear:comb}) as the basic building block.\footnote{Please refer to Appendix~\ref{appendix:incremental} for the implementation with adaptive conical combiners and incremental gradient descent.} 
In the case of MNIST, the reference signal $\myvec{y} \in \mathcal{R}^{10}$ is a ten-dimensional one-hot encoding of the correct digit, where the correct digit is represented by a one, and the other components are set to zero.
While the input signal $\myvec{x} \in \mathcal{R}^{28 \times 28}$ represents the corresponding input image without any additional preprocessing. 
The weights of each layer of adaptive combiners are obtained using their least squares solutions, \eqref{eq:cac:ls}.   
Building on the previous section, \figref{fig:mnist:cac} presents the overall model of applying CACs to the MNIST classification.

As a transformation of the original MNIST images, at each layer, we apply one of the two different transformations (cf. \figref{fig:mnist:cac}).  
The first transformation (solid lines in~\figref{fig:mnist:error}) flattens the image to $784$-dimensional vector and multiplies it by a random projection matrix. The result of matrix-vector multiplication is followed by $\mathrm{ReLU}$ activation function.  
The second transformation (dashed lines in~\figref{fig:mnist:error}) applies several random  convolutional filters $10 \times 10$ pixels each followed by max pooling with a stride of $2$, resulting in $81$ feature per filter (please see the corresponding implementation in Appendix~\ref{appendix:rand:conv}). 
For each layer, we generate new random projection matrices or new random convolutional filters.

\figref{fig:mnist:error} reports the prediction error on MNIST testing images for these transformations and the number layers in CACs ranging from $1$ to $20$.
For the convolutional CAC, we consider three different number of convolutions per layer $\{16,32,64\}$ while for the CAC with random projections we use the matching number of basis functions $\{1296,2592,5184\}$. 
For both transformations, increasing the number of layers in CACs reduces the error, which eventually plateaus. 
As the result, even the simplest transformation with random projections (solid lines) achieves the error rates below $2.0\%$ while using transformations with inductive bias for images (convolutions, dashed lines) can further reduce the error rates approaching $1.5\%$.

\section{Discussion}

This article introduced  Cascaded Adaptive Combiners (CAC), a biologically grounded computational approach that can serve as a building block for brain-like algorithms.
CAC operates through differential signaling, a metabolically efficient mechanism in which information is expressed relative to other signals. 
This property has important consequences for both learning and inference.

Learning within  a single CAC layer follows a greedy optimization strategy that depends only on the layer's feature vectors and the error terms from the previous layer (or the reference signal for the first layer; see \figref{fig:cac}C).
This strategy is conceptually related to the matching pursuit algorithm in signal processing~\cite{mallat1993matching}, where components are selected sequentially to minimize the error term.
Because weight updates rely solely on local information, CAC naturally supports online learning using algorithms such as recursive least squares and enables efficient offline learning via, e.g., regularized least squares.

This limited inter-layer dependency structure also supports  scalable and modular architectures.
Inference across all layers can be executed  in parallel within a single computational step, with the overall prediction obtained  as the sum of layer-wise outputs, \eqref{eq:prediction}.
Such inherent  parallelism makes CAC suitable for real-time applications and well aligned with specialized hardware such as graphics processing units, application-specific integrated circuits, and field-programmable gate arrays.

Beyond computational efficiency, CACs with adaptive conical combiners exhibit rich representational capabilities, allowing them to model and manipulate hierarchical structures.
In this formulation, error terms are computed within an algebra of convex cones rather than a conventional vector space, yet  the overall behavior of CAC remains analogous.
This algebra provides a promising framework for representing non-numeric conceptual entities~\citep{Nilsson.2023ipb}.
Open challenges include initializing a database of such entities and handling sparsity efficiently; both of which could further enhance hardware realizations of CAC.

Our MNIST digit classification experiment in \sectref{sec:combiners:mnist} illustrates potential of CAC and highlights a key challenge: selecting appropriate basis functions for input transformation. 
As shown in \figref{fig:mnist:error}, for a fixed number of basis functions, performance depends on the choice of input transformation.
Nevertheless, even randomized transformations yielded reasonable performance.
This observation mirrors findings in neuroscience, where both random~\cite{modi2020drosophila} and learned~\cite{olshausen1996emergence} basis functions coexist.
A single CAC layer with a random nonlinear expansion is functionally equivalent to a randomized feed-forward neural network~\cite{igelnik1995stochastic}.
Such randomized networks have demonstrated strong performance in tabular~\cite{HundredsClassifiers2014} in spatiotemporal~\cite{yan2024emerging} data processing, and connect naturally to the kernel methods literature~\cite{RCNNSsurvey}.
In particular, the random features approach~\cite{rahimi2007random} approximates well-known kernels such as the radial basis function kernel.
These conceptual links open new avenues for exploring design choices in selecting basis functions and nonlinearities for adaptive combiners.
For example, the adaptive filter (\sectref{sec:adaptive:filter}) resembles  certain recurrent randomized neural networks, such as architectures based on a single dynamical node with delayed feedback~\cite{appeltant2011information}. 
This connection suggests opportunities to explore CAC-based approaches for processing spatiotemporal data.

In summary, CAC provides a novel and versatile computational approach that combines biological accuracy, efficient learning, and expressive representational power. 
Future research should further investigate its theoretical properties, develop adaptive mechanisms for selecting basis functions, and pursue optimized implementations on parallel hardware to fully realize its potential.
We anticipate that this guide will encourage researchers and practitioners to explore CACs further and contribute to advancing this promising learning paradigm.

\section*{Acknowledgments}
This work was supported by the AFOSR under award number FA8655-25-1-7007.
DK acknowledges funding from the Swedish Research Council under the Starting Grant program (Grant No. 2025-05421), 

\section*{Declaration of Interests}
The authors declare no competing interests.

\bibliographystyle{splncs04}
\bibliography{ref}

\newpage

\appendix
\section*{Appendix: More elaborate program examples}
\renewcommand{\thesection}{\Alph{section}}

\section{Incremental solution}
\label{appendix:incremental}

This program below implements CAC for MNIST using incremental gradient descent rather than matrix pseudo-inverse. By constraining the weights to be non-negative, it realizes adaptive conical combiners, which is not possible with the pseudo-inverse approach.

In addition, the program uses PyTorch to offload as much computation as possible to the GPU when available. Its incremental formulation also makes it a suitable foundation for an online, streaming implementation.

\begin{python}[basicstyle=\scriptsize\ttfamily]
# -*- coding: utf-8 -*-
"""
Cascaded Adaptive Combiners - incremental version
=================================================
This code implements a cascaded adaptive combiner (CAC) network using
PyTorch. It trains and tests the network on the MNIST dataset of handwritten digits.
"""

import numpy as np
import torch as pt
import time

from torchvision import datasets
from torchvision.transforms import v2

ALMOST_ZERO = 1e-8      # To avoid division by zero
BATCH_SIZE = 10         # Mini-batch size

# ----- PyTorch special

# Check if GPU is available
if pt.cuda.is_available():
    print("GPU available")
else:
    print("Sorry, GPU unavailable, but you can still run PyTorch on the CPU")

# ----- MNIST data loading

# Define a transform to convert to array
transform = v2.Compose([
    v2.ToImage(),
    v2.ToDtype(pt.float32, scale=True)
    ])

# Load the training data
trainset = datasets.MNIST(root='./data', train=True, transform=transform,
                          download=True)
trainloader = pt.utils.data.DataLoader(trainset, batch_size=BATCH_SIZE,
                                       shuffle=True)

# Load the test data
testset = datasets.MNIST(root='./data', train=False, transform=transform)
testloader = pt.utils.data.DataLoader(testset, batch_size=BATCH_SIZE,
                                      shuffle=False)

# Sizes of x (features) and y (targets) input vectors (MNIST specific)
size_of_inputs = (28*28, len(trainset.classes))

# ----- Generate input data

# Decode target from vector
def ydecode(y):
    return np.argmax(y, axis=0)

# Define a generator for data to be input
def observations(train=True):
    """Yield successive mini-batches of MNIST inputs and one-hot labels.

    This is an infinite generator. Each iteration yields:
    - x : input matrix of shape (features, batch_size), moved to device
    - y : one-hot encoded class targets of shape (num_classes, batch_size)
    - labels : integer class labels (CPU tensor)

    Parameters
    ----------
    train : bool, optional
        If True, iterate over the training set; otherwise iterate over
        the test set.

    Yields
    ------
    (torch.Tensor, torch.Tensor, torch.Tensor)
        Tuple (x, y, labels) for each batch.
    """
        
    # Define loader depending on training or testing
    loader = trainloader if train else testloader

    while True:
        for images, labels in loader:
            batch_x = images.to(dev).view(images.size(0), -1).T  # (features, B)
            # One-hot encode labels directly, shape (num_classes, B)
            batch_y = pt.nn.functional.one_hot(labels,
                                               num_classes=size_of_inputs[1]) \
                        .to(dev).T.float()
            yield batch_x, batch_y, labels

# ----- Propagation (speed critical) all on GPU

def propagate(x, y, activelayer, rndmat, weights, cutoff, learningrate):
    """Propagate inputs through the CAC network and optionally update weights.

    The function applies:
    - Initial feature normalization
    - Cascaded random frame transforms
    - Per-layer residual updates
    - Optional weight updates for the active layer

    All computations are performed on the GPU, using in-place operations
    where possible to minimize memory usage.

    Parameters
    ----------
    x : torch.Tensor
        Input feature matrix of shape (features, batch_size).
    y : torch.Tensor
        Residual target matrix, modified in-place during propagation.
    activelayer : int
        Index (0-based) of the last layer to process.
    rndmat : torch.Tensor
        Random frame matrix of shape (features, features).
    weights : torch.Tensor
        Trainable coefficient tensor of shape
        (layers, num_classes, features).
    cutoff : float
        Lower bound applied via ReLU-like clamp before normalization.
    learningrate : float
        If > 0, perform weight updates for the active layer.

    Returns
    -------
    torch.Tensor
        The negative residual vector after propagation, shape
        (num_classes, batch_size).
    """
    
    # For initial layer
    x0 = x[:size_of_inputs[0],:]

    # Normalize initial feature vector , x0 = x0 / ||x0||
    x0 /= x0.norm(dim=0, keepdim=True).clamp_min_(ALMOST_ZERO)

    # Compute residual, y = y - W0 * x0
    y -= weights[0,:,:x0.shape[0]] @ x0
    
    # Remaining layers
    for layer in range(1, activelayer + 1):

        # Compute frame, x = ReLU(R * x)
        x = rndmat[:,:x.shape[0]].mm(x).clamp_min_(cutoff)

        # Normalize, x = x / ||x||
        x /= x.norm(dim=0, keepdim=True).clamp_min_(ALMOST_ZERO)

        # Compute residual, y = y - W * x
        y -= weights[layer].mm(x)

    # Special handling for first layer
    if activelayer == 0:
        x = x0

    # Adjust weights
    if learningrate > 0:
        # Iterative case, grad = (y * x^T) / batch_size
        grad = (y @ x.T) / y.shape[1]
        # Update weights for active layer, W = W + learningrate * grad
        w = weights[activelayer,:,:grad.shape[1]]
        w += learningrate * grad
        # Constrain weights to be non-negative, w = max(w, 0)
        w.clamp_min_(0)
        
    return -y

# ----- Training and testing

def learn(features=10000, layers=2, trainsize=60000,
          cutoff=-0.5, learningrate=1, testsize=10000, device='cuda'):
    """Train and evaluate a Cascaded Adaptive Combiner network on MNIST.

    The training proceeds layer-by-layer. For each layer:
    - Training batches are processed using `propagate` to refine weights.
    - RMS error over the training set is reported.
    - The model is then evaluated on the test set.

    Parameters
    ----------
    features : int
        Dimensionality of the random frame representation.
    layers : int
        Number of CAC layers to train sequentially.
    trainsize : int
        Number of training samples to use.
    cutoff : float
        Threshold used in nonlinear activation before normalization.
    learningrate : float
        Learning rate for weight updates. Set to 0 for frozen weights.
    testsize : int
        Number of test samples to evaluate.
    device : str
        Compute device ('cuda' or 'cpu').

    Side Effects
    ------------
    Initializes and updates global variables:
    - dev : torch.device
    - rndmat : torch.Tensor
    - weights : torch.Tensor

    Prints RMS training error and classification error per layer.
    """

    global weights
    global rndmat
    global dev
    
    # Set PyTorch device to 'cuda' or 'cpu'
    # We had better be explicit about devices because
    # pt.utils.data.DataLoader has issues with
    # pt.set_default_device().
    if device == 'cuda':
        dev = pt.device('cuda')
    else:
        dev = pt.device('cpu')
    
    # Initialize constant GPU data
    pt.random.manual_seed(1)
    rndmat = pt.normal(0, 1, size=(features, features), device=dev)
    
    # The crucial coefficient matrix lives in the GPU
    weights = pt.zeros((layers, size_of_inputs[1], features), device=dev)
    
    batch_size = trainloader.batch_size
    train_batches = trainsize // batch_size
    test_batches = testsize // batch_size

    # Iterate learning cycle over layers
    for activelayer in range(layers):
        
        # ----- Training
        
        # Start data generators for training
        traindatagenerator = observations(train = True)

        # Watch time and mean square error
        totalerror = 0
        t = time.time()
        
        # Iterate through training set
        for batch in range(train_batches):
        
            # Generate new observation and move to GPU
            x, y, _ = next(traindatagenerator)
        
            # Propagate through layers on GPU
            err = propagate(x, y.clone(), activelayer, rndmat, weights, cutoff,
                            learningrate)
            # Add error
            totalerror += err.pow(2).sum().item()
        
        print(f"\rRMS error after layer {activelayer}: "
              f"{np.sqrt(totalerror / trainsize):.5f}, "
              f"Time: {time.time() - t:.1f} s on {dev}")

        # ----- Testing
        
        # Start data generators for testing
        testdatagenerator = observations(train = False)
        
        errors = 0
        # Iterate through test set                
        for batch in range(test_batches):
            
            # Generate new observation and move to GPU
            x, y, y_true = next(testdatagenerator)
            y0 = pt.zeros_like(y)

            # Propagate through layers on GPU
            yapprox = propagate(x, y0, activelayer,
                                rndmat, weights, cutoff, learningrate = 0)
            yapprox /= yapprox.norm(dim=0, keepdim=True).clamp_min(ALMOST_ZERO)
            yapprox = yapprox.to('cpu').numpy()    
            y_true = y_true.to('cpu').numpy()    

            # Count errors
            errors += np.sum(y_true != ydecode(yapprox))

        print("\rTesting after layer", activelayer, "found", errors,
              "mistakes in", testsize, "classifications")

# ----- Sample run

if __name__ == "__main__":
    learn(layers=12, device='cuda')

\end{python}

The output is as follows:

\begin{python}[basicstyle=\scriptsize\ttfamily]
GPU available
RMS error after layer 0: 0.79169, Time: 14.9 s on cuda
Testing after layer 0 found 3108 mistakes in 10000 classifications
RMS error after layer 1: 0.59793, Time: 20.0 s on cuda
Testing after layer 1 found 1022 mistakes in 10000 classifications
RMS error after layer 2: 0.52614, Time: 28.1 s on cuda
Testing after layer 2 found 741 mistakes in 10000 classifications
RMS error after layer 3: 0.47258, Time: 40.4 s on cuda
Testing after layer 3 found 592 mistakes in 10000 classifications
RMS error after layer 4: 0.43747, Time: 51.5 s on cuda
Testing after layer 4 found 513 mistakes in 10000 classifications
RMS error after layer 5: 0.41475, Time: 63.9 s on cuda
Testing after layer 5 found 477 mistakes in 10000 classifications
RMS error after layer 6: 0.39862, Time: 75.1 s on cuda
Testing after layer 6 found 434 mistakes in 10000 classifications
RMS error after layer 7: 0.38634, Time: 86.8 s on cuda
Testing after layer 7 found 401 mistakes in 10000 classifications
RMS error after layer 8: 0.37673, Time: 98.1 s on cuda
Testing after layer 8 found 379 mistakes in 10000 classifications
RMS error after layer 9: 0.36885, Time: 109.9 s on cuda
Testing after layer 9 found 359 mistakes in 10000 classifications
RMS error after layer 10: 0.36224, Time: 121.2 s on cuda
Testing after layer 10 found 339 mistakes in 10000 classifications
RMS error after layer 11: 0.35696, Time: 132.7 s on cuda
Testing after layer 11 found 337 mistakes in 10000 classifications
\end{python}

\section{Solution using convolutions}
\label{appendix:rand:conv}
Similar to the program above, the next program uses the PyTorch package to perform as much processing as possible on the GPU, if available. It implements CAC for MNIST using the matrix pseudo-inverse and random convolutions for extracting feature vectors.

\begin{python}[basicstyle=\scriptsize\ttfamily]

#!/usr/bin/env python3
# -*- coding: utf-8 -*-
"""
Cascaded Adaptive Combiners - version with random convolutions
=================================================
This code implements a Cascaded Adaptive Combiner (CAC) network using PyTorch. 
It trains and tests the network on the MNIST dataset of handwritten digits.
"""

import torch
from torchvision import datasets
import time

############################## CAC FUNCTIONS ##################################
# ----- Propagation (speed critical) all on GPU (if available)
def lsq_solve(x_conv, y):
    """ Solve Least Squares system W X = Y for W """
    features = x_conv.size()[1]
    
    x_conv_prod = x_conv.T @ x_conv
    # Tichonov regularization below guarantees that the coefficient matrix is non-singular
    for k in range(features):
        x_conv_prod[k,k] += tichonov_lambda
    
    return torch.linalg.lstsq(x_conv_prod, (y @ x_conv).T, driver='gels').solution.T

def propagate(x, y, layers, testtarget = None, train = True):
    """ Training and testing core"""
    # Start timer
    t = time.time()
    
    size = x.shape[0]
    
    # Going via all layers
    for layer in range(layers):
        
        # Compute frame
        x_conv = pool(random_conv[layer](x))
        x_conv = torch.flatten(x_conv, start_dim=1, end_dim=-1) 

        # Normalize, ideally by the spectral norm for iterative solution
        x_conv /= torch.linalg.norm(x_conv, dim=1, keepdim=True)
        features = x_conv.size()[1]

        if train:            
            # Compute and store weights on CPU is slower if asynchronous
            w = lsq_solve(x_conv, y)
            weights[layer,:,:] = w.to(device='cpu', non_blocking=False)
        elif not train:
            # Load weights
            w = weights[layer].to(device=dev)
            
        # Compute residual
        y -= w[:,:features] @ x_conv.T           

        # Indicate progress and report results
        report_progress(t, y, layer, train, size, testtarget)
        
def report_progress(t, y, layer, train, size, target=None):
    """ Indicate progress and report results """
    if train:
        print("RMS error after layer %d, %.6f,"
              % (layer, torch.mean(torch.linalg.norm(y,dim=0))), end='')
    else: # Test
        # Decode target from vector
        errors = torch.sum(torch.argmax(torch.eye(ny, device=dev, dtype=torch.float32) @ -y, dim=0) != target, axis=0)
        print("Testing after layer %d, %d mistakes of %d classifications,"
              % (layer, errors, size), end='')
    print(" Time: %.1f s" % (time.time() - t), flush=True)    

# ----- Training and testing

def learn(num_bas, conv_size, layers, simul):
    """ Learn using cascaded adaptive combiners """
    
    global dev
    global tichonov_lambda
    global weights
    global random_conv
    global pool
    global X
    global Y    
    global Xts
    global Yts
    
    features = num_bas*(((npix-conv_size)//2)**2)    

    # Reset possibly set global arrays to free up memory
    weights = None
    
    # Initialize constant GPU data
    tichonov_lambda = torch.tensor(1.e-2, device=dev, dtype=torch.float32)

    # The crucial coefficient matrix lives in the CPU (not GPU!)
    weights = torch.zeros((layers, ny, features), device='cpu', dtype=torch.float32)
        
    # Create random convolutions for all layers of CAC
    random_conv = [torch.nn.Conv2d(1, num_bas, conv_size, stride=1, device = dev) for layer in range(layers)]
    pool = torch.nn.MaxPool2d(3, stride=2)

    for sim in range(simul):
        
        # ----- Training phase            
        # Propagate through layers on GPU
        propagate(X, torch.clone(Y.T), layers)  
    
        # ----- Testing phase
        # Create GPU buffers - physical storage order matters!
        Yts_pred = torch.zeros((testsize, ny), device=dev, dtype=torch.float32) 
    
        # Propagate through layers on GPU
        propagate(Xts, Yts_pred.T, layers, testtarget = Yts, train = False)    
        
        Yts_pred = None
            
        # Update random convolutions for the next simulation
        for layer in range(layers):
            random_conv[layer].reset_parameters()
###############################################################################

# ----- PyTorch special

# For reporoducibility set the seed
torch.random.manual_seed(1)

if torch.cuda.is_available():
    print("GPU available")
else:
    print("Sorry, GPU unavailable, but you can still run PyTorch on the CPU")

# Set pytorch device to 'cuda' or 'cpu'
dev = torch.device("cuda" if torch.cuda.is_available() else "cpu")    
###############################################################################

############################# LOAD MNIST DATA #################################
# Load the training data
trainset = datasets.MNIST(root='./data', train=True, download=True)
X = trainset.data.float().div(255.0).unsqueeze(1).to(dev)   # (60000, 1, 28 ,28)
Y = torch.nn.functional.one_hot(trainset.targets, num_classes=10).float().to(dev)

# Load the test data
testset = datasets.MNIST(root='./data', train=False, download=True)
Xts = testset.data.float().div(255.0).unsqueeze(1).to(dev)   # (10000, 1, 28 ,28)
Yts = testset.targets.float().to(dev)  

# Sizes of x (features) and y (targets) vectors
npix = 28
ny = 10     # Number of classes
testsize = Xts.shape[0]

############################# CAC PARAMETERS ################################## 
# Size of a convolutional filter
conv_size = 10
# Number of convolutional filter per layer
num_bas = 64 
# Number of layers in CAC
num_layers = 5
# Number of simulations to run
simul = 1
###############################################################################

# ----- Sample run  
learn(num_bas, conv_size, num_layers, simul)
\end{python}

The output is as follows:

\begin{python}[basicstyle=\scriptsize\ttfamily]
Sorry, GPU unavailable, but you can still run PyTorch on the CPU
RMS error after layer 0, 0.290419, Time: 10.2 s
RMS error after layer 1, 0.266948, Time: 18.8 s
RMS error after layer 2, 0.252380, Time: 27.6 s
RMS error after layer 3, 0.240673, Time: 36.1 s
RMS error after layer 4, 0.230567, Time: 44.3 s
Testing after layer 0, 222 mistakes of 10000 classifications, Time: 0.7 s
Testing after layer 1, 195 mistakes of 10000 classifications, Time: 1.5 s
Testing after layer 2, 192 mistakes of 10000 classifications, Time: 2.2 s
Testing after layer 3, 177 mistakes of 10000 classifications, Time: 2.9 s
Testing after layer 4, 171 mistakes of 10000 classifications, Time: 3.7 s
\end{python}

\end{document}